\documentclass[]{gtech}
\usepackage{amssymb}
\usepackage{multirow}
\usepackage{bigdelim}
\usepackage{todonotes}
\usepackage{longtable}
\usepackage{tabularray}
\usepackage{wrapfig}
\usepackage[most]{tcolorbox} 
\usepackage{xcolor}
\usepackage{url}

\renewcommand{\title}[1]{\newcommand{\titlelist}{{\huge\fontfamily{optimistic}\selectfont #1}}}

\newcommand{\model}{\texttt{Ming-Omni}}
\newcommand{\modellite}{\texttt{Ming-Lite-Omni}}
\newcommand{\modellinefirst}{\texttt{Ming-Lite}}
\newcommand{\modellinesecond}{\texttt{Omni}}

\definecolor{prompt}{HTML}{5f84e4}
\definecolor{img}{HTML}{820100}

\usepackage{arydshln}

\definecolor{CQColor}{rgb}{0.0,0.0,1.0} 

\definecolor{TSColor}{rgb}{0.5,0.0,0.8} 

\definecolor{CQRColor}{rgb}{1.0,0.0,1.0} 

\usepackage{colortbl}
\usepackage{amssymb}
\usepackage{pifont}
\usepackage{booktabs,multirow}
\usepackage{makecell}
\usepackage{tabulary}
\usepackage{fontawesome5}
\usepackage{bbding}
\usepackage{multicol}

\newlength\savewidth

\title{\textcolor[HTML]{0369ff}{Ming}-Omni: A Uni{f}{i}ed Multimodal Model for Perception and Generation}
\author[*]{Inclusion AI, Ant Group}

\contribution[*]{See Contributions section (Sec.~\ref{sec:contri}) for full author list.}

\abstract{\fontsize{11pt}{12pt} \textit{We propose Ming-Omni, a unified multimodal model capable of processing images, text, audio, and video, while demonstrating strong proficiency in both speech and image generation.
Ming-Omni employs dedicated encoders to extract tokens from different modalities, which are then processed by Ling, an MoE architecture equipped with newly proposed modality-specific routers.
This design enables a single model to efficiently process and fuse multimodal inputs within \textbf{a unified framework}, thereby facilitating diverse tasks without requiring separate models, task-specific fine-tuning, or structural redesign.
Importantly, Ming-Omni extends beyond conventional multimodal models by supporting audio and image generation. This is achieved through the integration of an advanced audio decoder for natural-sounding speech and Ming-Lite-Uni for high-quality image generation, which also allow the model to engage in context-aware chatting, perform text-to-speech conversion, and conduct versatile image editing. 
Our experimental results showcase Ming-Omni offers a powerful solution for unified perception and generation across all modalities.
Notably, our proposed Ming-Omni is the first open-source model we are aware of to match GPT-4o in modality support, and we release all code and model weights to encourage further research and development in the community.
}}

\date{May 21, 2025\vspace{-1mm}}
\gtechdata[Project Homepage]{\href{https://lucaria-academy.github.io/Ming-Omni/}{\underline{Link}}\vspace{-1mm}}
\gtechdata[Code]{\url{https://github.com/inclusionAI/Ming/tree/main}}

\begin{document}
\maketitle

\section{Introduction}
\label{sec:intro}

\label{sec:method}
\begin{figure}[t]
    \centering
    \includegraphics[width=1.0\linewidth]{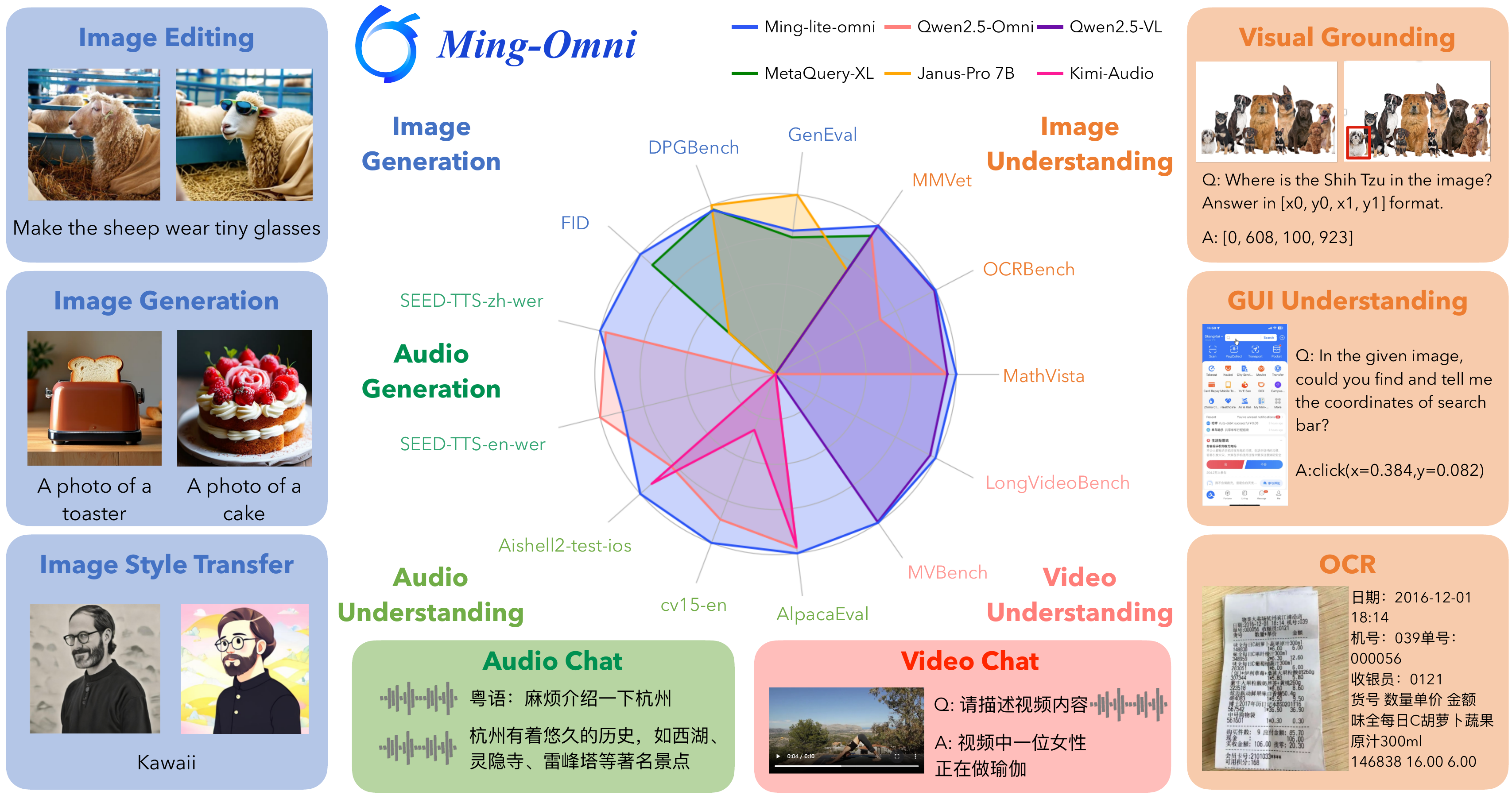}
    \caption{\model{} is a versatile, unified end-to-end model capable of processing images, text, video, and audio, and generating text, speech, and images. These capabilities enable the lite version of our model, \modellite{}, to support a broad range of tasks, including visual perception, audio-visual interaction, and image generation, among others.}
    \label{Ming_moe_uni}
    \vspace{-4mm}
\end{figure}

Humans effortlessly integrate visual and auditory cues to express ideas and generate vivid mental imagery from descriptions, supporting creativity, problem-solving, and communication as core aspects of intelligence.
The ultimate goal of Artificial General Intelligence (AGI) is to emulate this form of human-like multimodal intelligence, gradually evolving from a tool into a highly capable agent that enhances and liberates human productivity.
The recent advance of Large Language Models (LLMs), coupled with extensive training on vast multi-modal datasets, have catalyzed the emergence of strong perceptual capabilities in both vision~\citep{chen2024internvl,bai2025qwen2,team2025kimi} and audio~\citep{ding2025kimi,xu2025qwen2}, as well as generative capabilities in these two paradigms~\citep{huang2025step,ding2025kimi,chatgpt4o,tong2024metamorph,pan2025transfer}. 
%
%
%
%
Nevertheless, it remains challenging how to effectively blend both modalities in a single understanding model. Beyond comprehension tasks, another critical hurdle is integrating robust generation capabilities into these models to produce coherent, context-aware outputs across modalities while maintaining semantic consistency.

Despite these advancements, constructing a unified Omni-MLLM is hindered by representational disparities across modalities and divergent convergence rates during training.
To address these challenges, \model{} employs a language model with an integrated Mixture-of-Experts (MoE) architecture~\citep{ling}, where modality-specific routing is enabled through dedicated mechanisms for each type of token, allowing for tailored routing distributions.
%
Furthermore, following~\citep{guo2025m2}, we apply a stepwise balancing strategy during pre-training to mitigate cross-modal data imbalance, and employ a dynamically adaptive approach during instruction tuning to align training progress across modalities, leading to better convergence and model performance.
Through the optimization of model architecture and training strategies, \model{} achieves robust perception and understanding across multiple modalities.




Building on this robust perceptual foundation, we extend \model{} with an audio decoder and Ming-Lite-Uni~\citep{Ming-Lite-Uni}, enabling joint audio and image generation.
For speech generation, we address challenges like prosodic naturalness, real-time response, context awareness, and complex acoustic environments (\textit{e.g.}, dialects or accents) via two innovations: 1) \model{} uses Byte Pair Encoding (BPE) to improve prosody and reduce token frame rate by 35\%, boosting real-time performance. 2) A two-stage training strategy that prevents audio understanding and generation tasks from influencing each other, where the first stage focuses on understanding capabilities and the second stage concentrates on generation quality.
This also enables \model{} to balance efficiency and naturalness across diverse linguistic scenarios. 
Turning to visual generation, reconciling disparate visual feature representations is a fundamental challenge in unified multi-modal models. While existing methods like TokenFlow~\citep{qu2024tokenflow} and Janus~\citep{deepseek_janus} achieve impressive results, they compromise semantic fidelity due to pixel-centric optimization. In contrast, \model{} adopts a lightweight bridging framework that keeps the MLLM frozen and generates images progressively from coarse to fine using multi-scale learnable tokens guided by an alignment strategy.
A dedicated connector integrates latent representations produced by the MLLM with the diffusion decoder, leveraging its semantic understanding for image generation.

These architectural innovations empower \model{} to deliver exceptional cross-modal performance, as validated across image perception, audio-visual interaction, and image generation tasks. Specifically, in the image perception task, \model{} attained performance comparable to that of Qwen2.5-VL-7B~\citep{bai2025qwen25vltechnicalreport} by activating only 2.8B parameters.
%
\model{} delivers superior performance in end-to-end speech understanding and instruction following, surpassing Qwen2.5-Omni~\citep{xu2025qwen2} and Kimi-Audio~\citep{ding2025kimi}. It also supports native-resolution image generation, editing, and style transfer, achieving a GenEval score of 0.64, outperforming mainstream models such as SDXL~\citep{podell2023sdxl}. In terms of FID, \model{} reaches 4.85, setting a new SOTA across existing methods.

The key features of \model{} can be summarized as follows.
\begin{itemize}

\item \textbf{Unified Omni-Modality Perception:}
\model{}, built on Ling~\citep{ling}, an MoE architecture LLM, resolves task conflicts and ensures coherent integration of tokens from different modalities through modality-specific routers.

\item \textbf{Unified Perception and Generation:}
\model{} achieves unified understanding and generation, enabling the model to interpret multimodal instructions and user intent during generation, thereby improving generation quality and usability across multiple tasks.

\item \textbf{Innovative Generation Capabilities:} \model{} can perceive all modalities and generate high-quality text, real-time speech, and vivid images simultaneously, delivering exceptional cross-modal performance across diverse tasks including image perception, audio-visual interaction, and image generation.
\end{itemize}

\section{Approach}
\label{sec:method}
\begin{figure}[t]
    \centering
    \includegraphics[width=0.9\linewidth]{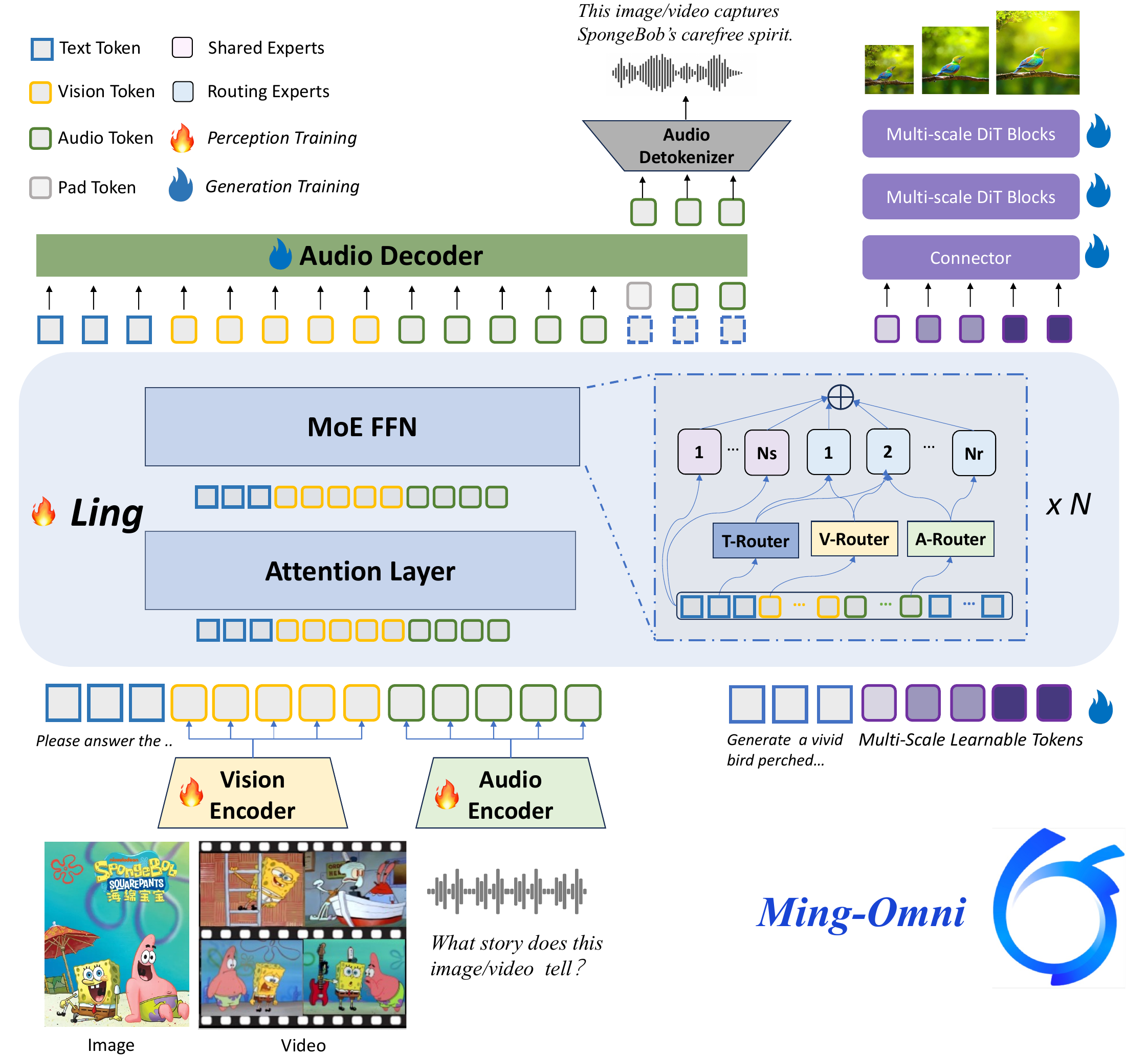}
    \vspace{-2mm}
    \caption{The overall framework of \model{}. \model{} extracts visual and audio tokens with dedicated encoders. These tokens are then combined with text tokens and processed through Ling (MoE architecture with modality-specific routers). Subsequently, it generates speech through an audio decoder and enables image generation via a diffusion model.}
    \label{Ming_moe_uni}
    \vspace{-2mm}
\end{figure}

As illustrated in Figure~\ref{Ming_moe_uni}, \model{} is a unified model capable of supporting inputs from multiple modalities, including images, audio, video, and text, while also facilitating speech generation and image generation. The overall training process of \model{} is divided into two distinct phases: perception training and generation training. During the perception training phase, the focus is on training the language interface Ling~\citep{ling} to effectively perceive and understand visual and audio tokens across different modalities. In the generation training phase, the emphasis shifts to training the audio decoder and the DiT module to enhance the model's generative capabilities. In the subsequent sections, we will explore how \model{} achieves unified understanding and generation across all modalities.

\subsection{Unified Understanding Across Modalities}

The cornerstone of \model{} lies in its state-of-the-art (SOTA) capability for comprehensive multimodal understanding. To achieve this, \model{} integrates the Qwen2.5~\citep{bai2025qwen25vltechnicalreport} vision backbone as its visual encoder, which supports arbitrary resolutions and demonstrates superior performance in both image and video processing tasks. Additionally, \model{} utilizes Whisper~\citep{radford2023robust} as its audio encoder, which has been proven to exhibit robust performance in ASR and speech understanding tasks. The embeddings generated by these encoders are projected to align with the dimensionality of the language model. These projected embeddings are then concatenated with the tokenized text inputs and fed into the language model Ling, which is based on a MoE architecture. Given the challenges of training an omni-modal large language model (omni-MLLM) due to the incongruence in representational spaces of different modalities and the disparity in convergence rates across modalities, we optimized the architecture of Ling by designing distinct routers for different modalities. This design enables tokens from each modality to be routed to specialized experts, thereby facilitating more precise and efficient routing of modality-specific information. To further address these challenges, we adopted a step-wise balance strategy during the pre-training stage and introduced a dynamic adaptive balance strategy during the instruction tuning stage, following~\citep{guo2025m2}. The dynamic adaptive balance strategy dynamically adjusts the loss weights based on the convergence rates of each modality, thereby mitigating conflicts between different modalities and ensuring optimal training progress across all modalities.

\subsection{Unified Speech Understanding and Generation}
\label{sec_2_speech_understanding}

In the \model{} framework, following Qwen-Audio~\citep{chu2024qwen2}, we adopt Whisper as the audio encoder for its strong capabilities in audio modeling, which are not limited to human speech. We employ a combination of a linear layer and a convolutional downsampling layer to transform the audio encoder output features to the latent space of the Ling language model.
To unleash the pre-existing world knowledge and the question answering capability in pre-trained language model in audio processing, a highly diverse audio data corpus is collected with metadata attributes (such as conversational or command scenarios, ambient environment, \textit{etc.}) labeled by an audio labeler (see section \ref{part_4_audio_data}). On top of that, we find it critical to incorporate these additional metadata attributes into the instruction prompts of the speech understanding tasks in an optional manner, thereby offering the model supplementary contextual cues to enhance its comprehension performance. Furthermore, we instruct the model to first predict the language identifier of the input audio before performing downstream tasks such as speech recognition. Our experiments show that this strategy significantly improves overall performance, especially in dialectal speech recognition. Overall, our training framework—which incorporates contextual and language information—delivers substantial improvements over conventional ASR-inspired paradigms.

To generate audio contents, we follow CosyVoice and connect an audio decoder to the output of the language model, where the audio decoder is an autoregressive architecture that generates discrete audio tokens extracted by an audio tokenizer. For the generation of audio contents in the paradigm of end-to-end multimodal LLMs (MLLMs), two key challenges are encountered for high-quality speech generation: (i) the gap in the sequence length between text and audio tokens in autoregressive modeling; and (ii) the difficulty in generating speech responses closely aligned with the input context containing various input modalities.

To address the difference in the sequence length between the text and audio tokens, instead of using the discrete audio tokens directly as the target for the audio decoder, we apply Byte Pair Encoding (BPE) to the discrete tokens extracted by the audio tokenizer. The BPE efficiently compresses the token length of the discrete tokens by 36\% (from 50Hz to approximately 32Hz), which effectively improves the training and inference efficiency. Since the BPE encoding process is entirely reversible, the efficiency gain is obtained without any loss on the quality of generated audio content. Further, similar to the BPE in the language domain, it also encourages learning of the combinatorial information within the audio content and thus enhances the prosodic performance.

To promote the relevance between the generated audio content and the input multi-modal context, we follow Qwen2.5-Omni~\citep{xu2025qwen2} and feed the hidden state of the original input from the MLLMs to the audio decoder. This encourages the audio decoder to capture paralinguistic information in the original input, such as emotions and environmental details. However, we find joint training of MLLMs and the audio decoder in Qwen2.5-Omni brings the difficulty in optimizing both understanding and generation tasks, as is observed in~\citep{shi2025balancingspeechunderstandinggeneration}. Hence, during the training of audio generation modules, we freeeze the entire MLLMs and only train the audio decoder using text-to-speech data and multi-modal context-aware triplet data. This preserves the multimodal understanding capabilities of the MLLMs while achieving strong speech generative performance.

\subsection{Unified Image Understanding and Generation}
A key challenge in unifying image understanding and generation lies in maintaining high generation quality without compromising visual understanding capabilities.
This balance is typically reflected in performance trade-offs across benchmarks from both domains. 
A critical obstacle is enabling the image generator to make effective use of the multimodal language model’s internal representations, which encode rich world knowledge but are not directly aligned with visual outputs.

In pursuit of this goal, many existing approaches attempt to model visual tokens within a shared feature space for both understanding and generation. 
However, achieving a meaningful balance often requires complex design choices, including architectural modifications, loss balancing, and multi-stage training strategies. 
Without such mechanisms, improvements in one task often come at the cost of the other. As a result, tokens optimized for understanding tend to degrade generation quality, while tokens tuned for generation may reduce understanding accuracy.
We consider this a challenging yet potentially promising direction. Our ongoing efforts in this path will be presented in detail in a future version of this work.

\begin{figure*}[t]
\centering
\includegraphics[width=0.9\textwidth]{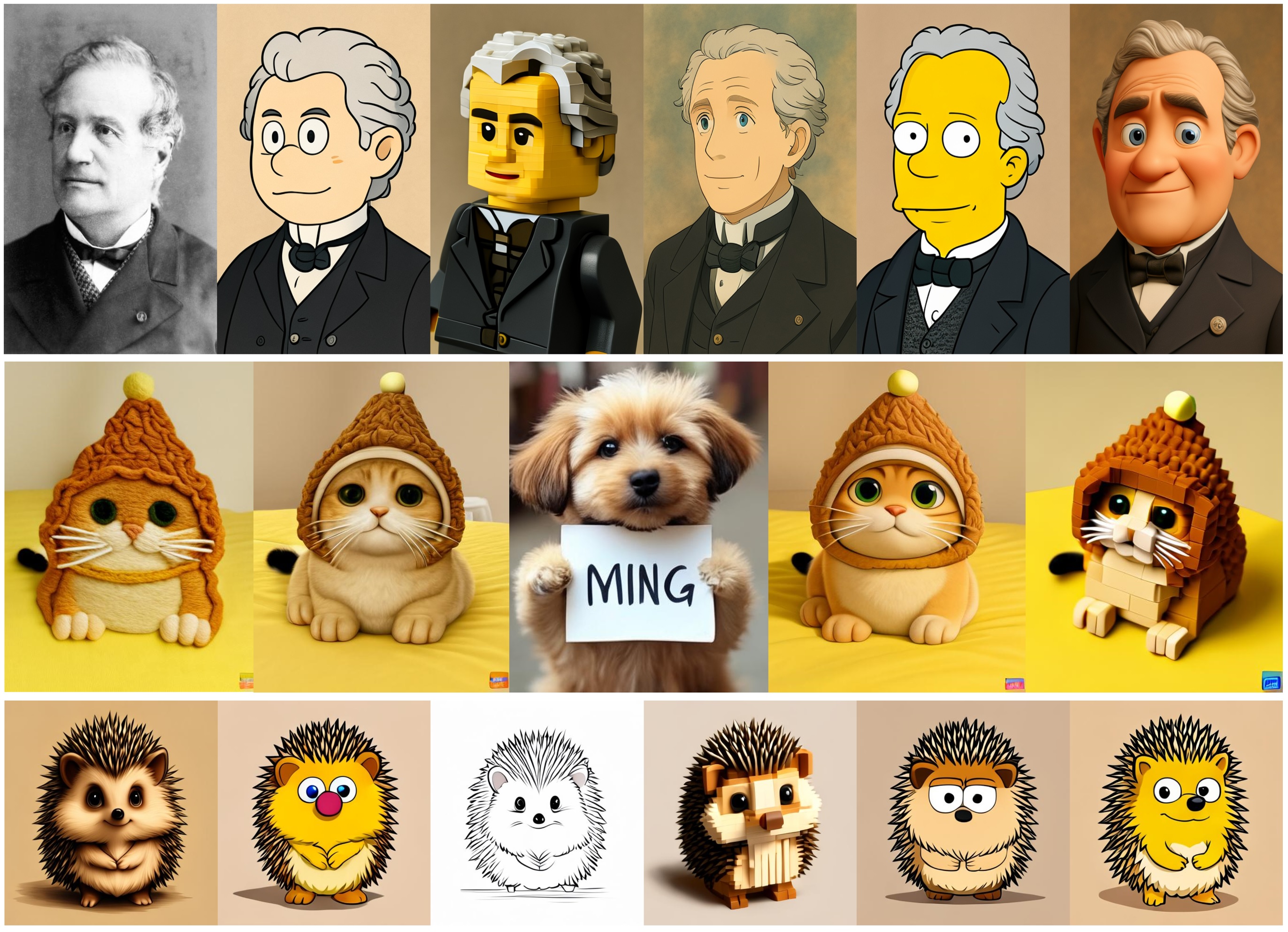}
\vspace{-4mm}
\caption{\centering Instruction based image style transfer results outputted by \model{}.}
\label{fig:styetransf}
\end{figure*}

At this released version, we propose a lightweight bridging approach which leverages multi-scale learnable tokens and multi-scale representation alignment. 
This design enables the model to adaptively benefit from tokens specialized for either understanding or generation. 
To avoid interference with image tokens for understanding, we introduce a dedicated image diffusion model during fine-tuning, ensuring high-quality generation while preserving semantic alignment and understanding performance.
Specifically, for understanding, we adopt the Qwen2.5-VL~\citep{bai2025qwen2} vision backbone with around 675 million parameters, retaining its architecture but retraining it jointly on images and videos. 
%
On the generation side and the bridging component, we introduce a novel multi-scale learnable token scheme, coupled with the representation-level alignment and the connector to enhance cross-task consistency. Details of key modules are described below.

\begin{figure*}[t]
\centering
\includegraphics[width=0.97\textwidth]{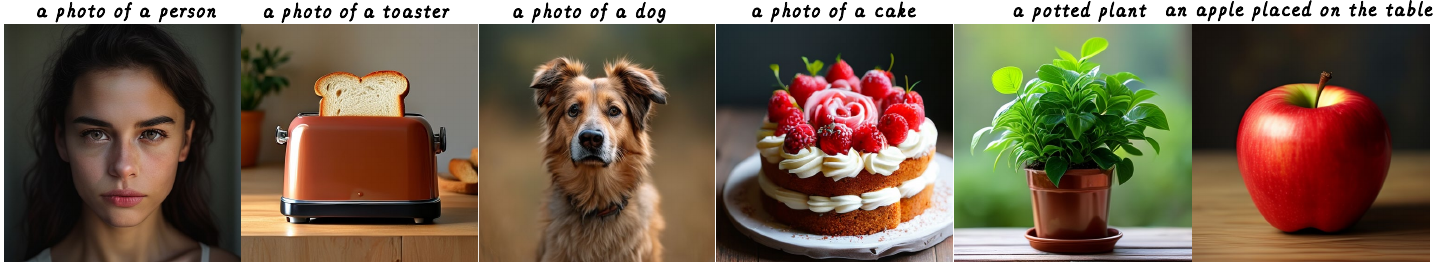}
\vspace{-4mm}
\caption{\centering Instruction based T2I results outputted by \model{}.}
\label{fig:imggen}
\end{figure*}

\begin{figure*}[t]
\centering
\includegraphics[width=0.97\textwidth]{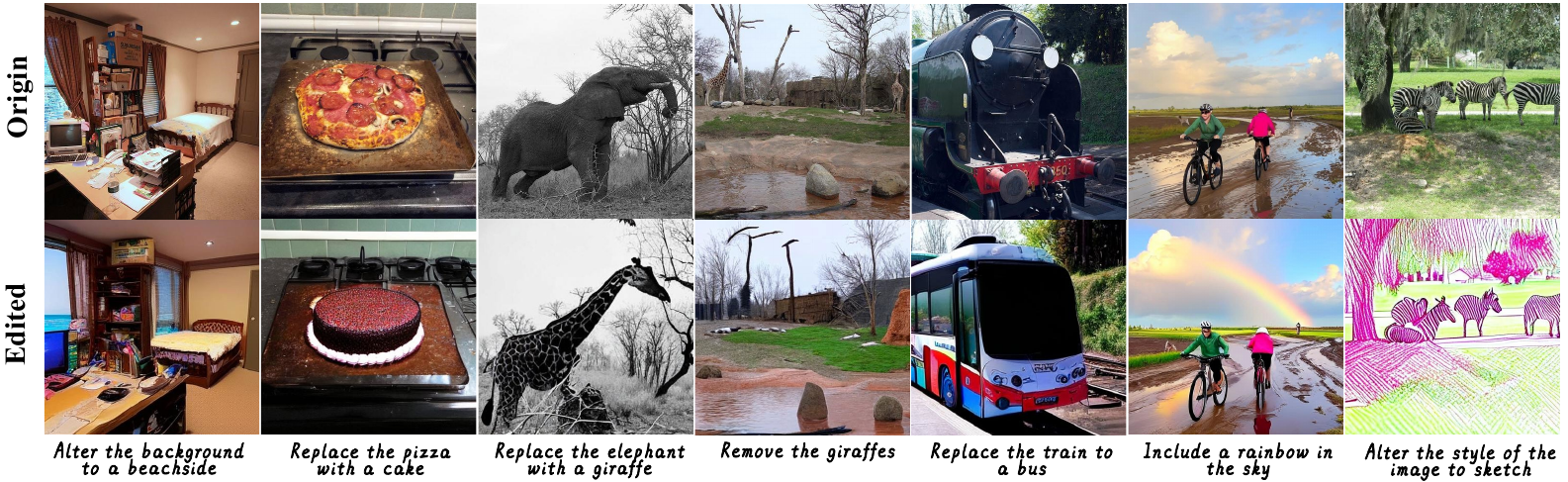}
\vspace{-4mm}
\caption{\centering Instruction based image editing results outputted by \model{}.}
\label{fig:imgediting}
\end{figure*}

\paragraph{\textbf{Multi-Scale Learnable Tokens Fusion and Processing}}
Given an input image $x$, we define a set of scales $\mathcal{S} = \{s_1, s_2, \dots, s_K\}$, where each $s_k$ corresponds to a spatial resolution, e.g., $s_k \in \{4\times4, 8\times8, 16\times16\}$. Each scale $s_k$ is associated with a dedicated set of learnable query tokens $Q_{s_k} \in \mathbb{R}^{N_{s_k} \times d}$, where $N_{s_k}$ is the number of tokens for scale $s_k$, and $d$ is the hidden dimension size. 
Each $Q_{s_k}$ is designed to capture information at different granularity levels, including global layout and color distribution, major objects and mid-level structures, and fine textures and detailed patterns. 
To preserve semantics at different spatial scales, we introduce explicit boundary markers by adding learnable start and end tokens around each scale's sequence. Each token is also assigned a scale-specific positional encoding derived from the spatial grid of that resolution. All scale-level sequences and their positional encodings are concatenated to form the input to the transformer encoder, which then processes this combined sequence to produce the final hidden representations.

\paragraph{\textbf{Multi-Scale Representation Alignment}} 
%
Beyond bridging with multi-scale learnable tokens, we further achieve implicit unification of understanding and generation through feature-level alignment.
Specifically, we align the intermediate hidden states from the DiT backbone with the final semantic representations by minimizing the mean squared error between them. This alignment loss encourages consistency between hierarchical representations and outputs through native-resolution optimization. 

As shown in Figure~\ref{fig:styetransf} /~\ref{fig:imggen} /~\ref{fig:imgediting}, \model{} supports both text-to-image generation and instruction-based image editing tasks, expanding its applicability across a broader range of creative and practical applications. For more details, please refer to Ming-Lite-Uni~\citep{Ming-Lite-Uni}.

\subsection{Overall Training Procedure}

The training procedure of \model{} comprises two stages: perception and generation training. The perception stage, consisting of pre-training, instruction tuning, and alignment tuning, is consistent with the M2-omni~\citep{guo2025m2} training pipeline. Specifically, both the pre-training and instruction tuning stages are divided into three sub-stages, with each sub-stage designed to incrementally incorporate additional tasks. After the perception training stage, we further expand \model{}'s multi-modal generation capabilities through the generation training stage. In particular, this stage consists of two parallel training tasks: text-to-speech and image generation, where we freeze the multimodal perception LLM and train additional generation components. For image generation training, only the added connector, multiscale learnable queries, and the DiT blocks are trained. For text-to-speech training, we train the audio decoder to support multi-modal context-aware dialogue.  Through the generation training stage, \model{} expands image and speech generation capabilities without compromising its multi-modal understanding capabilities.

\newpage

\section{Data Construction}
\label{sec:data}


\subsection{Data Overall}

\begin{figure}[bpth]
    \centering
    \includegraphics[width=1.0\linewidth]{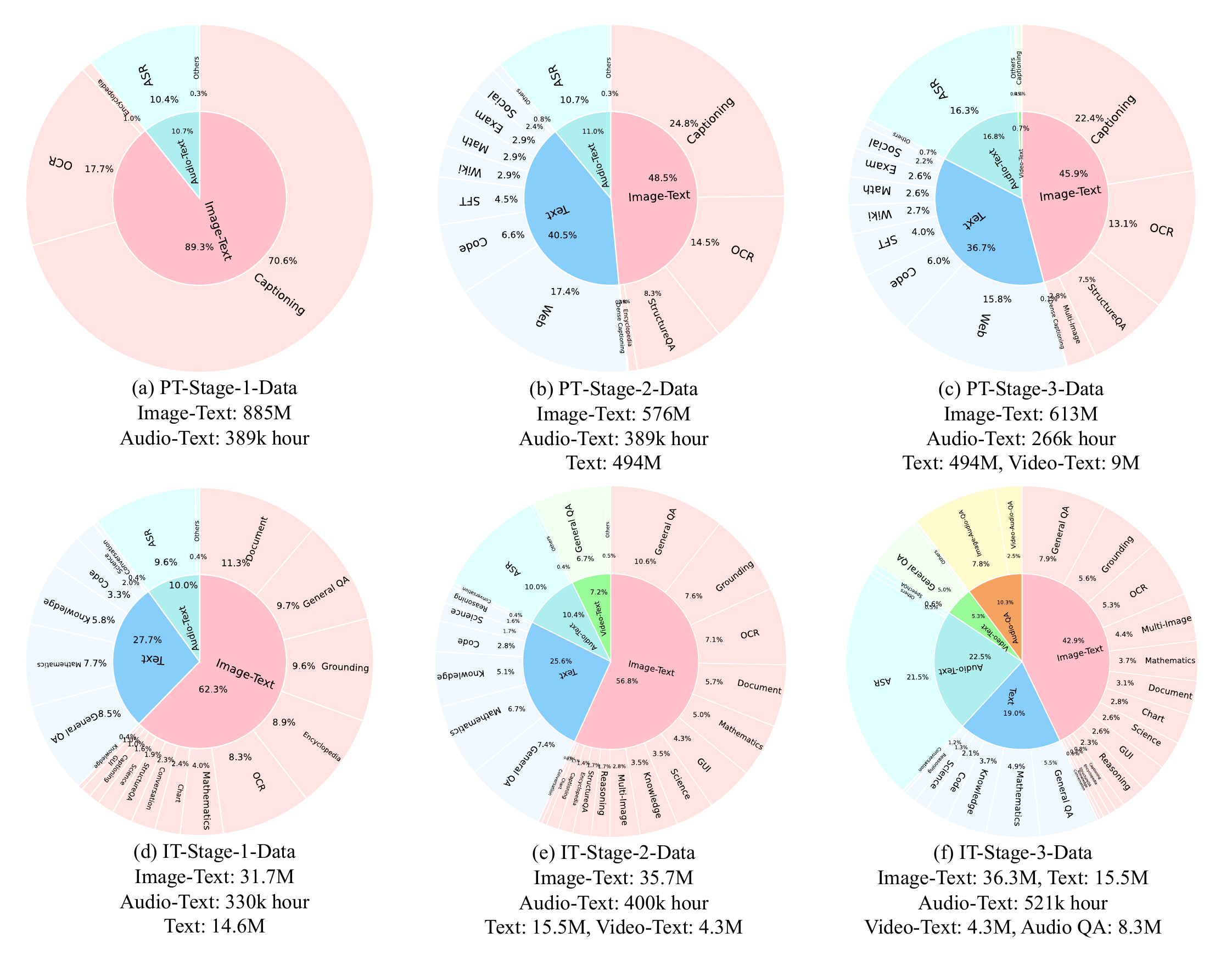}
    \caption{
    \textbf{Overview of the data configurations during pre-training and instruction tuning}. Note that the ``Others'' type of audio data includes AST, AAC, AAT, and SER data. The "Others" type of pre-training text data in (b) and (c) includes books, academic, news, and domain data. And the ``Others'' type of video-text data in (e) and (f) includes knowledge, captioning, and reasoning data. 
    }
    \label{fig-data-overall}
\end{figure}

We collect a large amount of training data covering different modalities and tasks. We built this diverse training data by collecting open source data and constructing multiple data production pipelines. The training objectives, modalities, and tasks of each training stage are different. Therefore, we construct different data configurations for each stage, as shown in Fig.~\ref{fig-data-overall}. Pre-training data, as shown in Fig.~\ref{fig-data-overall}~(a)-(c), is constructed adhering to two criteria: (1) aligning different modalities and (2) facilitating concept learning of world knowledge. Consequently, the resulting pre-training dataset exhibits diversity, aggregating multi-source data across various modalities. For instruction tuning, we aim to equip the model with the ability to address a broad spectrum of multimodal tasks. According to the tasks, we collect both open-source and in-house multimodal instruction tuning data for training, as shown in Fig.~\ref{fig-data-overall}~(d)-(f). The training data of alignment tuning, text-to-speech generation, and image generation are also constructed according to the corresponding tasks. The detailed data sources and data construction methods are further elaborated in this section.

\subsection{Image Data}
\label{part_3_1_image_data}
Image data serves as the cornerstone of our multi-modal data corpus. We integrate image understanding and generation datasets to enable our MLLM to derive unified perception-generation capabilities. Notably, we propose an iterative self-evolving framework that effectively improves the data quality and reduces the data volume; we also incorporate structured data and encyclopedia data that both empower our MLLM with fine-grained expert-level knowledge.



\subsubsection{Image Understanding Data}



\noindent\textbf{Caption Data}.
Caption data provides fundamental multi-modal alignment abilities. Our caption corpus is aggregated from a diverse collection of public datasets (\textit{e}.\textit{g}., Wukong~\citep{gu2022wukong}, Laion~\citep{laion}, DenseFusion~\citep{li2024densefusion}, ZERO-250M~\citep{zero-250m}, COYO-700M~\citep{byeon2022coyo}, \textit{etc.}) and in-house ones. Nevertheless, the presence of noisy, irrelevant, or incorrect image-caption pairs remains a critical challenge, leading to sub-optimal performance and hallucinatory MLLM responses. 
Inspired by DiverseEvol~\citep{wu2023self} and ASK-LLM~\citep{Sachdeva2024HowTT}, we propose an iterative self-evolving data refinement framework for caption corpus optimization.
Initially, we partition the full caption corpus $\mathcal{D}_{full}$ into ${T}$ non-overlapping split $\{{D}_{m}\}_{m=1}^{T}$, and randomly select one split to train the initial model $\mathbf{M}_{0}$, accompanied by an empty data pool $\mathcal{P}_{0}=\varnothing$ and a pre-defined filtering threshold $\tau$. At each iteration ${t}\in [1,{T}]$, we first employ the previous model $\mathbf{M}_{t-1}$ to evaluate all samples within another new split ${D}_{t}$. Samples whose inference scores exceed $\tau$ are grouped into ${D}_{t}^{high}$, while the others with low scores are discarded since they are often simple and trivial captions. Afterwards, we calculate quality scores on ${D}_{t}^{high}$ by designing tailored prompts to guide available MLLMs in assessing whether each sample is suitable for model training, only retaining verified high-quality samples to form the informative sub-split ${\ddot{D}}_{t}^{high}$. At the conclusion of each iteration ${t}$, we append ${\ddot{D}}_{t}^{high}$ to the evolving data pool $\mathcal{P}_{t}=\mathcal{P}_{t-1} \cup {\ddot{D}}_{t}^{high}$, and training another new model $\mathbf{M}_{t}$ based on $\mathcal{P}_{t}$ for the next iteration. After completing all $T$ iterations, we derive the refined caption corpus $\mathcal{P}_{T}=\{{\ddot{D}}_{t}^{high}\}_{t=1}^{T}$ to replace the initial set $\mathcal{D}_{full}$, thereby enhancing the quality and diversity of the caption corpus while significantly reducing the data volume.

\noindent\textbf{Structured Data}. Structured data enhances MLLM capabilities in addressing knowledge-intensive and information-seeking queries. Beyond conventional image understanding tasks that primarily require MLLM to recognize specific visual content like object detection and OCR, structured data presents challenges in querying fine-grained knowledge associated with specific visual entities. Inspired by InfoSeek~\citep{chen-etal-2023-pre-trained}, we develop an effective structured data synthesis pipeline to generate large-scale information-seeking QA pairs. Specifically, we first extract semantically significant entities from image content and their corresponding descriptions by leveraging multiple available MLLMs to perform a cross verification. We then employ a well-trained linking model to organize all extracted entities into a structured knowledge base. Afterwards, we synthesize information-seeking QA pairs by carefully designed prompts based on knowledge triplets, which embeds informative fine-grained knowledge.



\label{part_3_1_image_encyclopedia}
\noindent\textbf{Encyclopedia Data}. Encyclopedia data integrates advanced domain-specific expertise into MLLMs for expert-level comprehension and perception, \textit{e}.\textit{g}., identifying rare or endangered species via Latin binomial nomenclature. Our encyclopedia corpus spans 8 specific domains across biological categories (\textit{Plants and Animals}), cultural categories (\textit{Celebrities, Anime characters, and Artworks}), and daily-life categories (\textit{Ingredients, Dishes, and Vehicles}). To construct a high-quality expert-level corpus, we first collect a wide range of encyclopedia entities from academic databases and institutional websites. We then employ these entities as search queries to collect semantically relevant images via search engines. Afterwards, we develop a progressive encyclopedia data filtering scheme, including clip consistency validation, MLLM-based binary verification, and manual refinement.



\noindent\textbf{GUI Data}. GUI (Graphical User Interface) data enables MLLM to address complex real-world Android interaction tasks. Our GUI corpus is mainly constructed from three public datasets: AitW~\citep{AitW}, GUICourse~\citep{GUICourse}, and AndroidControl~\citep{li2024effectsdatascaleui}. Moreover, we leverage available MLLMs to optimize the reasoning process for each step within a human-like GUI interaction operation, which is subsequently reviewed by another MLLM to improve data quality.




\noindent\textbf{Reasoning Data}. Reasoning data activates the latent reasoning capabilities of MLLMs through the supervised long Chain-of-Thought (CoT) learning. Our reasoning corpus is mainly composed of two sources: textual reasoning data from Ling ~\citep{ling} and multi-modal reasoning data from R1-Onevision ~\citep{yang2025r1}. 




\label{part_3_1_image_preference}
\noindent\textbf{Preference Data}. Preference data optimizes MLLM responses through enhanced improved alignment with human-centric interaction patterns in the alignment tuning stage. Our preference corpus is primarily constructed from three sources: user-generated conversations in applications, search queries from websites, and high-quality instruction datasets. Specifically, we first retrieve relevant web images via search engines to complement text-only user-generated dialogues or queries. We then leverage available MLLMs to generate diverse specialized questions and their corresponding answers. Afterwards, we engage MLLMs and skilled human annotators to assess the quality of these QA pairs. Ultimately, we organize those high-quality positive samples and the other negative ones to construct the preference corpus, which comprises 41 subcategories across 9 primary domains.

\subsubsection{Image Generation Data}

Image generation data extends MLLM capabilities beyond conventional image understanding tasks. Following Ming-Lite-Uni~\citep{Ming-Lite-Uni}, our image generation corpus mainly comes from two sources: High-quality images collected from public image generation datasets (\textit{e}.\textit{g}., InstructPix2Pix-clip-filtered~\citep{brooks2023instructpix2pix}, SEED-Data-Edit-part2/3~\citep{ge2024seededit}, Ultra-edit~\citep{zhao2024ultraedit}, and \textit{etc.}); and image style transfer data sampled from StyleBooth and WikiArt. Readers can refer to Ming-Lite-Uni~\citep{Ming-Lite-Uni} for more construction details and data examples.



\subsection{Audio Data}
\label{part_4_audio_data}
As previously mentioned, the diversity of the audio data is critical for unleashing the pre-existing knowledge and capabilities in the pretrained language model to audio processing. Hence, we use a large amount of data from open-source datasets (detailed in Table~\ref{tab:audio_data}) as well as in-house datasets consisting of web data and synthetic data.
The web data are obtained from a carefully designed data filtering process.
\textbf{(i)} To begin with, we crawl a large set of audio data from the web based on a set of keywords that are expanded from handcrafted ones using hundreds of domain-specific lexical variations. \textbf{(ii)} VAD~\citep{gao2023funasrfundamentalendtoendspeech} is applied to obtain well-conditioned short audio clips. \textbf{(iii)} An audio labeler is trained iteratively, where the labeler is first trained with a high-quality dataset, before it is used to label the whole audio corpus. The labeled data are then further used to improve the precision of the audio labeler. 

Based on the short audio clips obtained from \textbf{(ii)} and the audio labeler from \textbf{(iii)}, we acquire a large number of high-quality audio clips with labels from different domains. 
Ablation studies are performed for an optimal data ratio among labels.
Empirical findings indicate that a larger number of English audio clips compared to Chinese audio clips results in significant improvements in English understanding, without degrading the ability to understand Chinese. Furthermore, dialect data constitutes only 2\% of the total data corpus, and the performance of dialect understanding plateaus as the ratio of dialect data increases.

\subsection{Video Data}

Video data enables MLLM to understand spatial-temporal content beyond sequences of static images. Our video corpus is mainly curated from open-source datasets (\textit{e}.\textit{g}., VideoGPT+~\citep{Maaz2024VideoGPT+}, Vript~\citep{yang2024vript}, Openvid-1M~\citep{nan2024openvid}, Youku-mPLUG-10M~\citep{xu2023youku}, \textit{etc.}) and public English / Chinese websites. Following LLaVA-Video~\citep{llava_video}, we adopt a hierarchical annotation pipeline to progressively generate dense video captions, open-ended question-answer pairs, and multi-choice question-answer pairs.





\subsection{Text Data}

Text data is essential for MLLM to maintain and improve language proficiency. Our text corpus is derived from two sources, \textit{i}.\textit{e}., Ling~\citep{ling} and M2-omni~\citep{guo2025m2}.

\begin{table*}[bthp]

\begin{minipage}[t]{1.0\linewidth}
        \caption{
            Performance of \modellite{} on \textbf{OpenCompass Image-Text Benchmarks} compared to leading MLLMs.
        }
	\small
	\vspace{-0.2cm}
	\centering
        \begin{tabular}{
            p{3.0cm}<{\centering}p{1.8cm}<{\centering}p{1.8cm}<{\centering}p{1.8cm}<{\centering}p{1.8cm}<{\centering}p{1.8cm}<{\centering}
        }
            \hline
		\multicolumn{1}{c|}{\multirow{2}{*}{Benchmark}} &
            \multicolumn{1}{c}{\modellinefirst{}} &
		\multicolumn{1}{c}{Qwen2.5} &
            \multicolumn{1}{c}{Qwen2.5VL} &
		\multicolumn{1}{c}{InternVL2.5} &
		\multicolumn{1}{c}{Gemma3}
		\\ 
		\multicolumn{1}{c|}{}&
		\multicolumn{1}{c}{\modellinesecond{}} &
            \multicolumn{1}{c}{Omni} &
		\multicolumn{1}{c}{7B-Instruct} &
		\multicolumn{1}{c}{8B-MPO} &
		\multicolumn{1}{c}{27B}
		\\ \hline
        \multicolumn{1}{c|}{AI2D} & 83.1 & 83.2 & 84.4 & \underline{\textbf{84.5}} & 81.4
		\\ 
        \multicolumn{1}{c|}{HallusionBench} & 55.0 & - & \underline{\textbf{55.8}} & 51.7 & 48.8
		\\ 
        \multicolumn{1}{c|}{MMBench-TEST-V11} & 80.8 & 81.8 & \underline{\textbf{82.8}} & 82.0 & 78.9
		\\ 
        \multicolumn{1}{c|}{MMMU} & 56.3 & 59.2 & 56.6 & 54.8 & \underline{\textbf{64.8}}
		\\ 
        \multicolumn{1}{c|}{MMStar} & 64.7 & 64.0 & \underline{\textbf{65.3}} & 65.2 & 59.6
		\\ 
        \multicolumn{1}{c|}{MMVet} & 71.3 & 66.8 & \underline{\textbf{71.6}} & 68.1 & 71.0
		\\ 
        \multicolumn{1}{c|}{MathVista} & \underline{\textbf{71.6}} & 67.9 & 68.1 & 67.9 & 67.6
		\\ 
        \multicolumn{1}{c|}{OCRBench} & \underline{\textbf{88.4}} & 57.8 & 87.8 & 88.2 & 75.3
		\\ \hline


        \label{table_1_ocnormal}
	\end{tabular}
\end{minipage}


\vspace{-0.2cm}

\begin{minipage}[t]{1.0\linewidth}
        \caption{
            Performance of \modellite{} on \textbf{PUBLIC Grounding Benchmarks} compared to leading MLLMs.
        }
	\small
	\vspace{-0.2cm}
	\centering
        \begin{tabular}{
            p{1.8cm}<{\centering}p{1.0cm}<{\centering}|p{2.0cm}<{\centering}p{2.0cm}<{\centering}p{2.0cm}<{\centering}p{2.0cm}<{\centering}p{2.0cm}<{\centering}
        }
            \hline
		\multicolumn{1}{c}{\multirow{2}{*}{Benchmark}} &
            \multicolumn{1}{c|}{\multirow{2}{*}{Split}} &
            \multicolumn{1}{c}{\modellinefirst{}} &
            \multicolumn{1}{c}{Qwen2.5-Omni} &
            \multicolumn{1}{c}{InternVL2.5} &
		\multicolumn{1}{c}{Grounding-DINO} &
		\multicolumn{1}{c}{Qwen2.5VL} 
		\\ 
		&
            &
		\multicolumn{1}{c}{\modellinesecond{}} &
            \multicolumn{1}{c}{7B} &
            \multicolumn{1}{c}{8B} &
            \multicolumn{1}{c}{Large} &
		\multicolumn{1}{c}{7B-Instruct} 
		
		\\ \hline
        & val & \underline{\textbf{90.6}} & 90.5 & 90.3 & \underline{\textbf{90.6}} & 90.0
		\\ 
        RefCOCO & testA & 93.1 & 93.5 & \underline{\textbf{94.5}} & 93.2 & 92.5
		\\ 
        & testB & 86.3 & 86.6 & 85.9 & \underline{\textbf{88.2}} & 85.4
		\\ \hline
        & val & \underline{\textbf{85.4}} & \underline{\textbf{85.4}} & 85.2 & 82.8 & 84.2
		\\ 
        RefCOCO+ & testA & 89.8 & 91.0 & \underline{\textbf{91.5}} & 89.0 & 89.1
		\\ 
        & testB & 79.2 & \underline{\textbf{79.3}} & 78.8 & 75.9 & 76.9
		\\ \hline
        \multicolumn{1}{c}{\multirow{2}{*}{RefCOCOg}} & val & 86.8 & \underline{\textbf{87.4}} & 86.7 & 86.1 & 87.2
		\\ 
        & test & 87.5 & \underline{\textbf{87.9}} & 87.6 & 87.0 & 87.2
		\\ 
        \hline
        \multicolumn{2}{c|}{Average} & 87.3 & \underline{\textbf{87.7}} & 87.6 & 86.6 & 86.6
		\\ \hline

        \label{table_4_ocobjdet}
	\end{tabular}
\end{minipage}


\vspace{-0.4cm}

\begin{minipage}[t]{0.50\linewidth}

        \captionsetup{font={small}}
        \caption{
            Performance of \modellite{} on \textbf{PUBLIC GUI Benchmarks} compared to leading MLLMs. The superscript ``$^{*}$'' denotes the reproduced results.
        }
        \small
	\vspace{-0.2cm}
	\centering
        \begin{tabular}{
            p{2.0cm}<{\centering}p{1.4cm}<{\centering}p{1.4cm}<{\centering}p{1.6cm}<{\centering}
        }
            \hline
		\multicolumn{1}{c|}{\multirow{2}{*}{Benchmark}} &
            \multicolumn{1}{c}{\modellinefirst{}} &
            \multicolumn{1}{c}{InternVL3} &
		\multicolumn{1}{c}{Qwen2.5VL}
		\\ 
            \multicolumn{1}{c|}{} &
		\multicolumn{1}{c}{\modellinesecond{}} &
            \multicolumn{1}{c}{8B} &
		\multicolumn{1}{c}{7B-Instruct}
		\\ \hline
        \multicolumn{1}{c|}{ScreenSpot} & \underline{\textbf{82.1}} & 79.5 & 78.9$^{*}$
		\\ 
        \multicolumn{1}{c|}{ScreenSpot-V2} & \underline{\textbf{84.1}} & 81.4 & -
		\\ 
        \multicolumn{1}{c|}{AITZ(EM)} & \underline{\textbf{66.6}} & - & 57.6$^{*}$
		\\ 
        \hline

        \label{table_3_ocgui}
	\end{tabular}


    \end{minipage}
\begin{minipage}[t]{0.04\linewidth}
\begin{tabular}{
        p{0.2cm}<{\centering}p{0.2cm}<{\centering}
    }
     &  
\end{tabular}
\end{minipage}
\begin{minipage}[t]{0.46\linewidth}

        \captionsetup{font={small}}
        \caption{
            Performance of \modellite{} on \textbf{PUBLIC  Information-Seeking Benchmarks} compared to leading MLLMs.
        }
        \small
	\vspace{-0.2cm}
	\centering
        \begin{tabular}{
            p{2.2cm}<{\centering}p{1.4cm}<{\centering}p{0.8cm}<{\centering}p{1.6cm}<{\centering}
        }
            \hline
		\multicolumn{1}{c|}{Benchmark} &
            \multicolumn{1}{c}{\modellinefirst{}} &
		\multicolumn{1}{c}{PaLI-X} &
		\multicolumn{1}{c}{Qwen2.5VL}
		\\ 
		\multicolumn{1}{c|}{(InfoSeek)} &
		\multicolumn{1}{c}{\modellinesecond{}} &
		\multicolumn{1}{c}{} &
            \multicolumn{1}{c}{32B}
		\\ \hline
        \multicolumn{1}{c|}{H-mean} & \underline{\textbf{27.7}} & 22.1 & 19.4
		\\ 
        \multicolumn{1}{c|}{Unseen-question} & \underline{\textbf{30.4}} & 23.5 & 20.6
		\\ 
        \multicolumn{1}{c|}{Unseen-entity} & \underline{\textbf{25.4}} & 20.8 & 18.3
		\\ 
        \hline
        
        \label{table_7_knowledge}
	\end{tabular}

    \end{minipage}

\vspace{-0.1cm}


\begin{minipage}[t]{1.0\linewidth}
        \caption{
            Performance of \modellite{} on \textbf{Text-to-Image Generation Benchmarks} compared to leading models. ``\textit{Gen.}'' denotes models for pure image generation, while ``\textit{Uni.}'' denotes models capable of both image understanding and generation. Note that the global best performance is highlighted by an \underline{underline}, and the local best result in ``\textit{Gen.}'' or ``\textit{Uni.}'' is marked with \textbf{bold}.
        }
	\small
	\vspace{-0.2cm}
	\centering
        \begin{tabular}{
            p{0.6cm}<{\centering}p{2.4cm}<{\centering}|p{0.8cm}<{\centering}p{0.8cm}<{\centering}p{0.8cm}<{\centering}p{0.8cm}<{\centering}p{0.8cm}<{\centering}p{0.8cm}<{\centering}p{0.8cm}<{\centering}<{\centering}|p{1.4cm}<{\centering}|p{0.5cm}
        }
            \hline
		\multicolumn{1}{c}{\multirow{2}{*}{Type}} &
            \multicolumn{1}{c|}{\multirow{2}{*}{Model}} &\multicolumn{7}{c|}{GenEval} &
             \multicolumn{1}{c|}{\multirow{2}{*}{DPG-Bench}} & \multicolumn{1}{c}{\multirow{2}{*}{FID$\downarrow$}}\\ 
             \cline{3-9}
        &&\multicolumn{1}{c}{1-Obj.} &\multicolumn{1}{c}{2-Obj.} &
		\multicolumn{1}{c}{Count} &
		\multicolumn{1}{c}{Colors} &
		\multicolumn{1}{c}{Posit.} &
		\multicolumn{1}{c}{Color.} &
            \multicolumn{1}{c|}{AVG} &&\\
        \hline
        & LlamaGen & 0.71 & 0.34 & 0.21 & 0.58 & 0.07 & 0.04 & 0.32 &- &-
		\\ 
        & LDM & 0.92 & 0.29 & 0.23 & 0.70 & 0.02 & 0.05 & 0.37 &- &-
		\\ 
        & SDv1.5 &  0.97 & 0.38 & 0.35 & 0.76 & 0.04 & 0.06 & 0.43 &- &-
		\\ 
        & PixArt-$\alpha$ &  0.98 & 0.50 & 0.44 & 0.80 & 0.08 & 0.07 & 0.48 & -&-
		\\ 
        \multicolumn{1}{c}{\textit{Gen.}} & SDv2.1 &  0.98 & 0.51 & 0.44 & 0.85 & 0.07 & 0.17 & 0.50 & 68.09 &26.96
		\\ 
        & Emu3-Gen & 0.98 & 0.71 & 0.34 & 0.81 & 0.17 & 0.21 & 0.54 & 80.60 & -
		\\ 
        & SDXL & 0.98 & 0.74 & 0.39 & 0.85 & 0.15 & 0.23 & 0.55 &74.65 & 8.76
		\\ 
        & DALL-E 3 & 0.96 & 0.87 & 0.47 & 0.83 & \textbf{0.43} & 0.45 & 0.67 & -&-
		\\ 
        & SD3-Medium  & \underline{\textbf{0.99}} & \underline{\textbf{0.94}} & \underline{\textbf{0.72}} & \underline{\textbf{0.89}} & 0.33 & \underline{\textbf{0.60}} & \textbf{0.74} &- &-
		\\ 
        \hline
        & LWM &  0.93 & 0.41 & 0.46 & 0.79 & 0.09 & 0.15 & 0.47 &- &-
		\\ 
        & SEED-X & 0.97 & 0.58 & 0.26 & 0.80 & 0.19 & 0.14 & 0.49 &- &-
		\\ 
        & Show-o &  0.95 & 0.52 & 0.49 & 0.82 & 0.11 & 0.28 & 0.53 & -&-
		\\ 
        \multicolumn{1}{c}{\textit{Uni.}} & TokenFlow-XL &  0.95 & 0.60 & 0.41 & 0.81 & 0.16 & 0.24 & 0.55 & -&-
		\\ 
        & Janus & 0.97 & 0.68 & 0.30 & \textbf{0.84} & \textbf{\underline{0.46}} & 0.42 & 0.61 & 79.68 & 10.10
		\\ 
        & JanusFlow &  - & - & - & - & - & - & 0.63 & 80.09 & 9.51
		\\ 
        & JanusPro-7B &  - & - & - & - & - & - & \textbf{\underline{0.80}} & \textbf{\underline{84.19}} & 13.48 \\
        \cdashline{2-11}
        & \modellite{} & \underline{\textbf{0.99}} & \textbf{0.77} & \textbf{0.68} & 0.78 & 0.31 & 0.29 & 0.64 & 81.72 & \textbf{\underline{4.85}}
		\\ 
        \hline
        
        \label{table_g_imggen}
	\end{tabular}

        \vspace{0.4cm}
\end{minipage}

\end{table*}






\begin{table*}[t]
    \small
    \begin{minipage}[t]{0.50\linewidth}

        \captionsetup{font={small}}
        \caption{
            Performance of \modellite{} on \textbf{PUBLIC OCR Benchmarks} compared to leading MLLMs. ``$\cdot$/$\cdot$'' denotes results under English (\textit{En}) and Chinese (\textit{Zh}) testing conditions as ``\textit{En}/\textit{Zh}''.
        }
        \small
	\vspace{-0.2cm}
	\centering
        \setlength\tabcolsep{0.1pt}%
        \begin{tabular}{
            p{1.6cm}<{\centering}p{1.6cm}<{\centering}p{1.4cm}<{\centering}p{1.4cm}<{\centering}p{1.6cm}<{\centering}
        }
            \hline
		\multicolumn{1}{c|}{\multirow{2}{*}{Benchmark}} &
            \multicolumn{1}{c}{\modellinefirst{}} &
            \multicolumn{1}{c}{InternVL3} &
            \multicolumn{1}{c}{Gemma3} &
		\multicolumn{1}{c}{Qwen2.5VL}
		\\ 
            \multicolumn{1}{c|}{} &
		\multicolumn{1}{c}{\modellinesecond{}} &
            \multicolumn{1}{c}{8B} &
            \multicolumn{1}{c}{27B} &
		\multicolumn{1}{c}{7B-Instruct}
		\\ \hline
        \multicolumn{1}{c|}{ChartQA} & 85.1 & 86.6 & 83.4 & \underline{\textbf{87.2}}
		\\ 
        \multicolumn{1}{c|}{DocVQA} & 93.0 & 92.7 & 89.5 & \underline{\textbf{95.6}}
		\\ 
        \multicolumn{1}{c|}{OCRBench-v2} & 53.3/52.0 & - & - & \underline{\textbf{56.3}}/\underline{\textbf{57.2}}
		\\ 
        \multicolumn{1}{c|}{TextVQA} & 82.8 & 80.2 & 83.2 & \underline{\textbf{85.1}}
            \\ 
        \multicolumn{1}{c|}{OmniDocBench$\downarrow$} & 34.0/\underline{\textbf{34.4}} & - & - & \underline{\textbf{30.8}}/39.8
		\\ \hline

        \label{table_3_ococr}
	\end{tabular}

    \end{minipage}
\begin{minipage}[t]{0.04\linewidth}
\begin{tabular}{
        p{0.2cm}<{\centering}p{0.2cm}<{\centering}
    }
     &  
\end{tabular}
\end{minipage}
\begin{minipage}[t]{0.46\linewidth}

        \captionsetup{font={small}}
        \caption{
            Performance of \modellite{} on \textbf{IN-HOUSE Encyclopedia Benchmarks} compared to leading MLLMs.
        }
        \small
	\vspace{-0.2cm}
	\centering
        \begin{tabular}{
            p{1.4cm}<{\centering}p{1.4cm}<{\centering}p{1.4cm}<{\centering}p{0.8cm}<{\centering}p{0.6cm}<{\centering}
        }
            \hline
		\multicolumn{1}{c|}{Benchmark} &
            \multicolumn{1}{c}{\modellinefirst{}} &
		\multicolumn{1}{c}{Qwen2.5VL} &
		\multicolumn{1}{c}{Ovis2} &
		\multicolumn{1}{c}{Ola}
		\\ 
		\multicolumn{1}{c|}{(In-house)} &
		\multicolumn{1}{c}{\modellinesecond{}} &
            \multicolumn{1}{c}{7B-Instruct} &
            \multicolumn{1}{c}{8B} &
            \multicolumn{1}{c}{7B}
		\\ \hline
        \multicolumn{1}{c|}{Plant} & \underline{\textbf{54.96}} & 47.85 & 37.83 & 36.33
		\\ 
        \multicolumn{1}{c|}{Animal} & \underline{\textbf{56.70}} & 50.85 & 45.49 & 40.38
		\\ 
        \multicolumn{1}{c|}{Vehicle} & 41.91 & 42.29 & \underline{\textbf{46.01}} & 39.05
		\\ 
        \multicolumn{1}{c|}{Ingredient} & \underline{\textbf{62.28}} & 54.09 & 54.72 & 53.88
		\\ 
        \multicolumn{1}{c|}{Dish} & \underline{\textbf{44.30}} & 39.07 & 42.56 & 42.56
		\\ 
        \hline
        \multicolumn{1}{c|}{Average} & \underline{\textbf{52.03}} & 46.83 & 45.32 & 42.44
		\\ \hline
        \label{table_6_eyes}
	\end{tabular}

    \end{minipage}

    \vspace{-0.4cm}

\begin{minipage}[t]{0.54\linewidth}
     
        \captionsetup{font={small}}
        \caption{
            Performance of \modellite{} on \textbf{PUBLIC Video Understanding Benchmarks} compared to leading MLLMs. All models are evaluated based on 128 uniformly sampled frames.
        }
	\small
	\vspace{-0.2cm}
	\centering
        \begin{tabular}{
            p{2.2cm}<{\centering}p{1.4cm}<{\centering}p{1.6cm}<{\centering}p{1.6cm}<{\centering}
        }
            \hline
		\multicolumn{1}{c|}{\multirow{2}{*}{Benchmark}} &
            \multicolumn{1}{c}{\modellinefirst{}} &
		\multicolumn{1}{c}{Qwen2.5VL} &
		\multicolumn{1}{c}{LLaVA-One} 
		\\ 
		\multicolumn{1}{c|}{} &
		\multicolumn{1}{c}{\modellinesecond{}} &
		\multicolumn{1}{c}{7B-Instruct} &
            \multicolumn{1}{c}{Vision-7B} 
		\\ \hline
        \multicolumn{1}{c|}{MVBench} & \underline{\textbf{67.7}} & 67.4 & 56.7
		\\ 
        \multicolumn{1}{c|}{VideoMME} & 67.0 & \underline{\textbf{67.3}} & 58.2
		\\ 
        \multicolumn{1}{c|}{VideoMMMU} & 46.3 & \underline{\textbf{47.4}} & 33.9
		\\ 
        \multicolumn{1}{c|}{LongVideoBench} & \underline{\textbf{56.6}} & 54.7 & 50.5
		\\ 
        \hline
        \multicolumn{1}{c|}{Average} & \underline{\textbf{59.4}} & 59.2 & 49.8
		\\ \hline
        
        \label{table_11_video}
	\end{tabular}


    \end{minipage}
\begin{minipage}[t]{0.04\linewidth}
\begin{tabular}{
        p{0.2cm}<{\centering}p{0.2cm}<{\centering}
    }
     &  
\end{tabular}
\end{minipage}
\begin{minipage}[t]{0.40\linewidth}

        \captionsetup{font={small}}
        \caption{
            Performance of \modellite{} on \textbf{IN-HOUSE Human Preference Benchmarks}.
        }
        \small
	\vspace{-0.2cm}
	\centering
        \begin{tabular}{
            p{2.4cm}<{\centering}p{1.6cm}<{\centering}p{1.6cm}<{\centering}
        }
            \hline
		\multicolumn{1}{c|}{Benchmark} &
            \multicolumn{1}{c}{\modellinefirst{}} &
		\multicolumn{1}{c}{Qwen2.5VL}
		\\ 
		\multicolumn{1}{c|}{(In-house)} &
		\multicolumn{1}{c}{\modellinesecond{}} &
            \multicolumn{1}{c}{7B-Instruct}
		\\ \hline
        \multicolumn{1}{c|}{Relevance} & \underline{\textbf{4.479}} & 4.308
		\\ 
        \multicolumn{1}{c|}{Fluency} & \underline{\textbf{4.907}} & 4.765
		\\ 
        \multicolumn{1}{c|}{Richness} & 3.498 & \underline{\textbf{3.828}}
		\\ 
        \multicolumn{1}{c|}{Appropriateness} & \underline{\textbf{4.740}} & 4.727
		\\ 
        \multicolumn{1}{c|}{Correctness} & \underline{\textbf{3.856}} & 3.741
		\\ 
        \hline
        \multicolumn{1}{c|}{Average Score} & \underline{\textbf{4.296}} & 4.274
		\\ \hline
        \label{table_5_prefence}
	\end{tabular}

        \vspace{-0.2cm}

    \end{minipage}

\vspace{-0.4cm}
 
\end{table*}






\begin{table*}[bthp]

        \captionsetup{font={small}}
        \caption{
            Performance of \modellite{} on \textbf{PUBLIC and IN-HOUSE Audio Understanding Benchmarks} compared to leading Models, including five different dimensions.
        }
        \small
	\vspace{-0.2cm}
	\centering
        \begin{tabular}{
            p{1.6cm}<{\centering}p{8.0cm}<{\centering}p{1.6cm}<{\centering}p{1.6cm}<{\centering}p{1.6cm}<{\centering}p{1.6cm}<{\centering}
        }
            \hline
            \multicolumn{1}{c}{\multirow{2}{*}{Type}} &
		\multicolumn{1}{c|}{\multirow{2}{*}{Benchmark}} &
            \multicolumn{1}{c}{\modellinefirst{}} &
		\multicolumn{1}{c}{Qwen2.5} &
            \multicolumn{1}{c}{Qwen2} &
            \multicolumn{1}{c}{Kimi}
		\\
            &
            \multicolumn{1}{c|}{} &
		\multicolumn{1}{c}{\modellinesecond{}} &
		\multicolumn{1}{c}{Omni} &
		\multicolumn{1}{c}{Audio} &
		\multicolumn{1}{c}{Audio}
		\\ \hline
        & \multicolumn{1}{c|}{Aishell1 $\downarrow$} & 1.47 & 1.18 & 1.53 & \underline{\textbf{0.60}}
		\\ 
        & \multicolumn{1}{c|}{Aishell2-test-android $\downarrow$} & \underline{\textbf{2.55}} & 2.75 & 2.92 & 2.64
		\\ 
        \textit{PUBLIC} & \multicolumn{1}{c|}{Aishell2-test-ios $\downarrow$} & \underline{\textbf{2.52}} & 2.63 & 2.92 & 2.56
		\\ 
        \textit{Chinese} & \multicolumn{1}{c|}{Cv15-zh $\downarrow$} & 6.31 & \underline{\textbf{5.20}} & 6.90 & 7.21
		\\ 
        \textit{Benchmarks} & \multicolumn{1}{c|}{Fleurs-zh $\downarrow$} & 2.96 & 3.00 & 7.50 & \underline{\textbf{2.69}}
		\\ 
        & \multicolumn{1}{c|}{Wenetspeech-testmeeting $\downarrow$} & 5.95 & \underline{\textbf{5.90}} & 7.16 & 6.28
            \\ 
        & \multicolumn{1}{c|}{Wenetspeech-testnet $\downarrow$} & 5.46 & 7.70 & 8.42 & \underline{\textbf{5.37}}
            \\ 
            \cdashline{2-6}
        & \multicolumn{1}{c|}{Average (Chinese) $\downarrow$} & \underline{\textbf{3.89}} & 4.05 & 5.34 & 3.91
            \\ 
            \hline
        & \multicolumn{1}{c|}{Llibrispeech-test-clean $\downarrow$} & 1.44 & 1.80 & 1.60 & \underline{\textbf{1.28}}
		\\ 
        & \multicolumn{1}{c|}{Librispeech-test-other $\downarrow$} & 2.80 & 3.40 & 3.60 & \underline{\textbf{2.42}}
		\\ 
        \textit{PUBLIC} & \multicolumn{1}{c|}{Multilingual-librispeech $\downarrow$} & \underline{\textbf{4.15}} & 7.56 & 5.40 & 5.88
		\\ 
        \textit{English} & \multicolumn{1}{c|}{Cv15-en $\downarrow$} & \underline{\textbf{6.89}} & 7.60 & 8.60 & 10.31
		\\ 
        \textit{Benchmarks} & \multicolumn{1}{c|}{Fleurs-en $\downarrow$} & \underline{\textbf{3.39}} & 4.10 & 6.90 & 4.44
		\\ 
        & \multicolumn{1}{c|}{Voxpopuli-v1.0-en $\downarrow$} & \underline{\textbf{5.80}} & \underline{\textbf{5.80}} & 6.84 & 7.97
		\\ 
            \cdashline{2-6}
        & \multicolumn{1}{c|}{Average (English) $\downarrow$} & \underline{\textbf{4.08}} & 5.04 & 5.49 & 5.38
            \\ 
        \hline
        & \multicolumn{1}{c|}{Hunan $\downarrow$}  & \underline{\textbf{7.88}} & 29.31 & 25.88 & 31.93
		\\ 
        \textit{IN-HOUSE} & \multicolumn{1}{c|}{Minnan $\downarrow$}  & \underline{\textbf{13.84}} & 53.43 & 123.78 & 80.28
		\\ 
        \textit{Dialect} & \multicolumn{1}{c|}{Guangyue $\downarrow$}  & \underline{\textbf{4.36}} & 10.39 & 7.59 & 41.49
		\\ 
        \textit{Benchmarks} & \multicolumn{1}{c|}{Chuanyu $\downarrow$}  & \underline{\textbf{4.33}} & 7.61 & 7.77 & 6.69
		\\ 
        & \multicolumn{1}{c|}{Shanghai $\downarrow$}  & \underline{\textbf{10.49}} & 32.05 & 31.73 & 60.64
		\\ \cdashline{2-6}
        & \multicolumn{1}{c|}{Chat $\downarrow$}  & \underline{\textbf{2.34}} & 3.68 & 4.29 & 2.96
		\\ 
        \textit{IN-HOUSE} & \multicolumn{1}{c|}{Government $\downarrow$}  & \underline{\textbf{1.77}} & 2.23 & 2.70 & 2.03
		\\ 
        \textit{Domain} & \multicolumn{1}{c|}{Health $\downarrow$}  & 3.31 & 4.02 & 4.18 & \underline{\textbf{2.38}}
		\\ 
        \textit{Benchmarks} & \multicolumn{1}{c|}{Knowledge $\downarrow$}  & 3.69 & 3.17 & 3.33 & \underline{\textbf{1.98}}
		\\ 
        & \multicolumn{1}{c|}{Local-live $\downarrow$}  & 2.44 & \underline{\textbf{2.03}} & 2.34 & 2.05
		\\ 
            \cdashline{1-6}
        \multicolumn{2}{c|}{Average (All IN-HOUSE) $\downarrow$} & \underline{\textbf{5.45}} & 14.79 & 21.36 & 23.24 
		\\ \hline

        \label{table_8_audioasr}
	\end{tabular}

        \vspace{0.0cm}

        \captionsetup{font={small}}
        \caption{
            Performance of \modellite{} on \textbf{PUBLIC Audio Question-Answering Benchmarks} compared to leading models. ``\textit{Audio.}'' denotes models specialized in pure audio understanding tasks, while ``\textit{Omni.}'' denotes models capable of multi-modal perception and generation capabilities beyond audio-centric tasks. Note that the global best performance is highlighted by an \underline{underline}, and the local best result in ``\textit{Audio.}'' and ``\textit{Omni.}'' is marked with \textbf{bold}.
        }
        \small
	\vspace{-0.2cm}
	\centering
        \begin{tabular}{
            p{0.6cm}<{\centering}p{2.2cm}<{\centering}p{1.4cm}<{\centering}p{1.6cm}<{\centering}|p{1.4cm}<{\centering}|p{1.0cm}<{\centering}p{1.8cm}<{\centering}|p{1.4cm}<{\centering}|p{1.4cm}<{\centering}
        }
            \hline
            \multicolumn{1}{c}{\multirow{2}{*}{Type}} &
		\multicolumn{1}{c|}{\multirow{2}{*}{Models}} &
            \multicolumn{2}{c|}{Open-ended QA} &
            \multicolumn{1}{c|}{Knowledge} &
            \multicolumn{2}{c|}{Multi-Choice QA} &
            \multicolumn{1}{c|}{Instruction} &
            \multicolumn{1}{c}{Safety}
		\\ 
            &
            \multicolumn{1}{c|}{} &
		\multicolumn{1}{c}{AlpacaEval} &
            \multicolumn{1}{c|}{CommonEval} &
            \multicolumn{1}{c|}{SD-QA} &
            \multicolumn{1}{c}{MMSU} &
            \multicolumn{1}{c|}{OpenBookQA} &
            \multicolumn{1}{c|}{IFEval} &
            \multicolumn{1}{c}{AdvBench}
		\\ \hline
        & \multicolumn{1}{c|}{Step-Audio-chat} & 3.99 & 2.99 & 46.84 & 31.87 & 29.19 & \underline{\textbf{65.77}} & 86.73 \\
        & \multicolumn{1}{c|}{Qwen2-Audio-chat} & 3.69 & 3.40 & 35.35 & 35.43 & 49.01 & 22.57 & 98.85 \\
        \textit{Audio.} & \multicolumn{1}{c|}{Baichuan-Audio} & 4.00 & 3.39 & 49.64 & 48.80 & 63.30 & 41.32 & 86.73 \\
        & \multicolumn{1}{c|}{GLM-4-Voice} & 4.06 & 3.48 & 43.31 & 40.11 & 52.97 & 24.91 & 88.08 \\
        & \multicolumn{1}{c|}{Kimi-Audio} & \textbf{4.46} & \textbf{3.97} & \underline{\textbf{63.12}} & \underline{\textbf{62.17}} & \underline{\textbf{83.52}} & 61.10 & \underline{\textbf{100.00}} \\
        \hline
        & \multicolumn{1}{c|}{Megrez-3B-Omni} & 3.50 & 2.95 & 25.95 & 27.03 & 28.35 & 25.71 & 87.69 \\
        & \multicolumn{1}{c|}{DiVA} & 3.67 & 3.54 & 57.05 & 25.76 & 25.49 & 39.15 & 98.27 \\
        
        \multicolumn{1}{c}{\multirow{2}{*}{\textit{Omni.}}} & \multicolumn{1}{c|}{Qwen2.5-Omni} & 4.49 & 3.93 & 55.71 & \textbf{61.32} & \textbf{81.10} & 52.87 & \textbf{99.42} \\
        & \multicolumn{1}{c|}{Baichuan-Omni-1.5} & 4.50 & 4.05 & 43.40 & 57.25 & 74.51 & 54.54 & 97.31 \\
        & \multicolumn{1}{c|}{MiniCPM-o} & 4.42 & \underline{\textbf{4.15}} & 50.72 & 54.78 & 78.02 & 49.25 & 97.69 \\
        \cdashline{2-9}
        & \multicolumn{1}{c|}{\modellite{}} & \underline{\textbf{4.63}} & 4.06 & \textbf{58.84} & 47.53 & 61.98 & \textbf{58.36} & 99.04 \\
        \hline

        \label{table_9_audioqa}
	\end{tabular}

        \vspace{0.0cm}

    
        \captionsetup{font={small}}
        \caption{
            Performance of \modellite{} on \textbf{PUBLIC Text-to-Speech Benchmarks} compared to leading MLLMs. \modellite{}-context denotes the model trained with multi-modal context-aware audio triplet data.
        }
        \small
	\vspace{-0.2cm}
	\centering
        \begin{tabular}{
            p{0.6cm}<{\centering}p{2.2cm}<{\centering}|p{1.4cm}<{\centering}p{1.8cm}<{\centering}p{1.0cm}<{\centering}p{1.4cm}<{\centering}p{0.8cm}<{\centering}p{0.8cm}<{\centering}p{1.6cm}<{\centering}p{1.2cm}<{\centering}
        }
            \hline
            \multicolumn{1}{c}{\multirow{2}{*}{Type}} &
		\multicolumn{1}{c|}{Benchmark} &
            \multicolumn{1}{c}{\modellinefirst{}} &
            \multicolumn{1}{c}{\modellinefirst{}} &
            \multicolumn{1}{c}{Seed} &
            \multicolumn{1}{c}{MaskGCT} &
            \multicolumn{1}{c}{E2} &
            \multicolumn{1}{c}{F5} &
            \multicolumn{1}{c}{CosyVoice2} &
		\multicolumn{1}{c}{Qwen2.5}
		\\ 
		&
            \multicolumn{1}{c|}{(Seed-TTS-Eval)} &
		\multicolumn{1}{c}{\modellinesecond{}} &
            \multicolumn{1}{c}{\modellinesecond{}-context} &
		\multicolumn{1}{c}{TTS} &
            \multicolumn{1}{c}{} &
            \multicolumn{1}{c}{TTS} &
            \multicolumn{1}{c}{TTS} &
            \multicolumn{1}{c}{} &
            \multicolumn{1}{c}{Omni}
		\\ \hline
        \multicolumn{1}{c}{\multirow{2}{*}{\textit{Chinese}}}  & Zh-wer $\downarrow$ & 1.69 & 1.98 & \underline{\textbf{1.11}} & 2.27 & 1.97 & 1.56 & 1.45 & 1.70
		\\ 
         & Zh-sim $\uparrow$ & 0.68 & 0.68 & \underline{\textbf{0.80}} & 0.77 & 0.73 & 0.74 & 0.75 & 0.75
		\\ \hline
        \multicolumn{1}{c}{\multirow{2}{*}{\textit{English}}} & En-wer $\downarrow$ & 4.31 & 5.10 & 2.24 & 2.62 & 2.19 & \underline{\textbf{1.83}} & 2.57 & 2.72
		\\ 
        & En-sim $\uparrow$ & 0.51 & 0.51 & \underline{\textbf{0.76}} & 0.71 & 0.71 & 0.65 & 0.65 & 0.63
		\\ \hline

        \label{table_10_audiotts}
	\end{tabular}
 
\end{table*}


\begin{figure*}[bthp]
    \centering
    \includegraphics[width=0.98\linewidth]{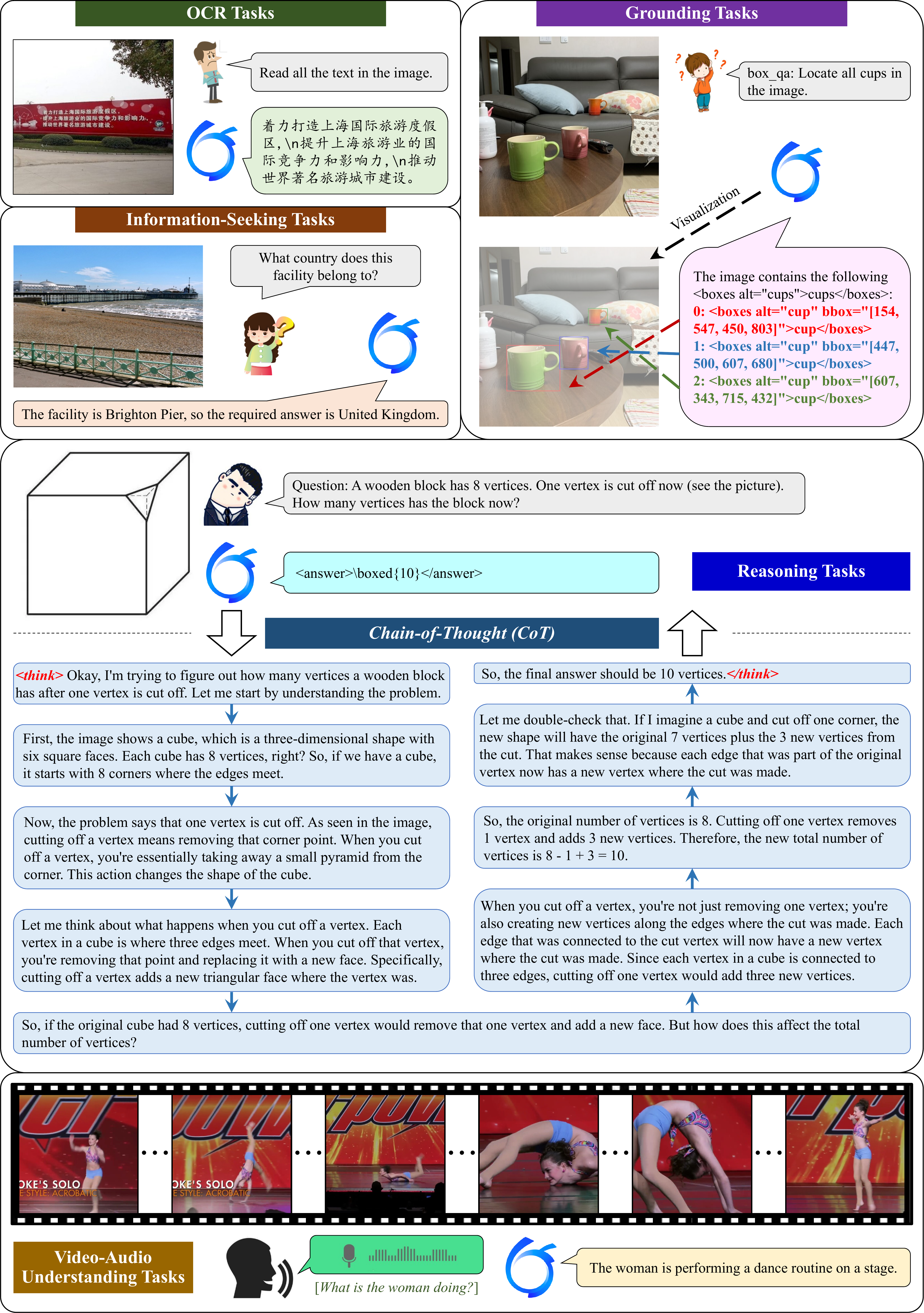}
    \caption{
        Visualization results of OCR, Grounding, Information-Seeking, Reasoning, and Video-Audio Understanding tasks.
    }
    \label{part-6-figure-under-all}
\end{figure*}







\begin{figure*}[bthp]
    \centering
    \includegraphics[width=0.98\linewidth]{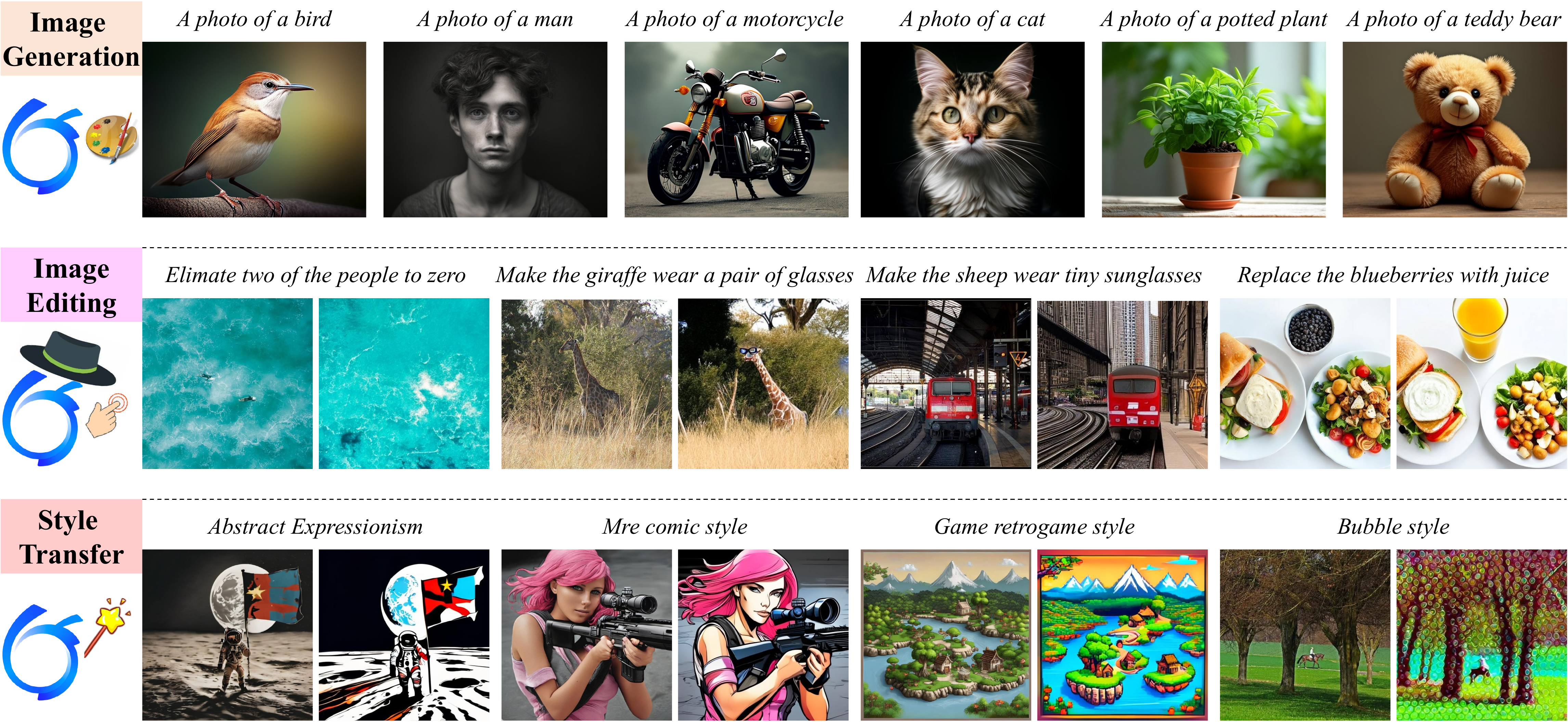}
    \caption{
        Visualization results of {Text $\rightarrow$ Image} tasks, including image generation task, image editing task, and style transfer task.
    }
    \label{part-5-fig-4-vis-gen}
\end{figure*}







\section{Evaluation}

We benchmark \modellite{}, a light version of \model{}, which is derived from Ling-lite and features 2.8 billion activated parameters, and compare it against leading SoTA MLLMs with under 10B parameters.

\subsection{Public Benchmarks}
\label{sec:B_5_1_public}

 As shown in Table \ref{table_1_ocnormal}$\sim$\ref{table_10_audiotts}, our holistic assessment covers more than 50 rigorously curated public benchmarks across the following five distinct multi-modal dimensions: 

\noindent\textbf{Image $\rightarrow$ Text (Understanding)}. Our evaluation of the image-to-text understanding capabilities primarily encompasses the following six tasks: 1) Fundamental image understanding capabilities evaluated on OpenCompass image-text comprehensive benchmarks~\citep{2023opencompass}. 2) Image Reasoning capabilities evaluated on OpenCompass image-text reasoning benchmarks~\citep{2023opencompass}. 3) Image Grounding capabilities evaluated on RefCOCO~\citep{refcoco}, RefCOCO+~\citep{refcoco}, and  RefCOCOg~\citep{refcocog}. 4) OCR capabilities evaluated on ChartQA~\citep{masry-etal-2022-chartqa}, DocVQA~\citep{docvqa}, OCRBenchV2~\citep{ocrbenchv2}, OmniDocBench~\citep{omnidocbench}, and TextVQA-VAL~\citep{Eval_textvqa}. 5) GUI capabilities evaluated on ScreenSpot~\citep{screenspot}, ScreenSpot-V2~\citep{wu2024atlas}, and AITZ(EM)~\citep{zhang2024android}. And 6) knowledge-based Question Answering capabilities evaluated on InfoSeek~\citep{chen2023can}.

\noindent\textbf{Text $\rightarrow$ Image (Generation)}. We incorporate text-to-image generation capabilities to enable our MLLM with unified perception-generation abilities, which are evaluated on GenEval~\citep{ghosh2024geneval}, DPG-Bench~\citep{hu2024ella}, and FID.

\noindent\textbf{Audio $\rightarrow$ Text (Understanding)}. Our evaluation of the audio-to-text understanding capabilities mainly includes the following two tasks: 1) Fundamental audio understanding capabilities evaluated on a broad range of public benchmarks, including public Chinese benchmarks like Aishell1~\citep{AISHELL1} and Wenetspeech~\citep{WenetSpeech}, and public English benchmarks like Librispeech~\citep{Librispeech} and Voxpopuli~\citep{wang2021voxpopuli}. And 2) audio question-answering capabilities evaluated on various benchmarks across five specific tasks, such as AlpacaEval and CommonEval from VoiceBench~\citep{chen2024voicebenchbenchmarkingllmbasedvoice} for open-ended QA tasks, and SD-QA for knowledge-based QA tasks.

\noindent\textbf{Text $\rightarrow$ Audio (Generation)}. We incorporate text-to-audio generation capabilities to enable our MLLM with unified audio perception-generation abilities, which are evaluated on Seed-TTS-Eval~\citep{Seed_TTS}.

\noindent\textbf{Video $\rightarrow$ Text (Understanding)}. Our evaluation of the video-to-text understanding capabilities contains the following four benchmarks: MVBench~\citep{li2024mvbench}, VideoMME~\citep{fu2024video}, VideoMMMU~\citep{hu2025video}, and LongVideoBench~\citep{wu2024longvideobench}.


\subsection{In-house Benchmarks}
\label{sec:B_5_2_inhouse}

In addition to public benchmarks, we also establish three in-house benchmarks to comprehensively evaluate multiple capabilities of MLLMs, including:

\noindent\textbf{Encyclopedia Benchmark}. We build an encyclopedia benchmark to adequately evaluate the expert-level comprehension capabilities of MLLMs. Following the encyclopedia data construction pipeline in Sec. \ref{part_3_1_image_encyclopedia}, we first extract representative image-text pairs, and engage skilled human annotators to verify their correctness and manually write four negative answer candidates. Subsequently, another group of experienced human annotators review the human-generated multi-choice questions, ensuring that four incorrect options are semantically different from the correct answer, while sharing some visual similarities that could confuse less capable models. 
The final encyclopedia benchmark comprises a total of 29,924 samples across five encyclopedia categories: Plants (7,225 cases), Animals (7,192 cases), Ingredients (7,123 cases), Dishes (6,530 cases), and Vehicles (1,854 cases).

\noindent\textbf{Human Preference Benchmark}. We construct an in-house human preference benchmark to evaluate the human-centric interaction patterns exhibited in MLLM responses. Following the preference data construction pipeline in Sec. \ref{part_3_1_image_preference}, all testing samples are reviewed by skilled human annotators to ensure the constructed benchmark is clean and reliable. Our benchmark contains a total of 1,025 samples. Inspired by SuperClue-V, we evaluate the quality of MLLM responses from five key dimensions: Relevance, Fluency, Information Richness, Format Appropriateness, and Correctness.



\noindent\textbf{Multi-Dialect and Multi-Domain Audio Understanding Benchmark}. We extend audio understanding benchmarks by incorporating two established testing sets to evaluate the capabilities of MLLMs in multi-dialect and multi-domain settings. Specifically, we collect audios from real users across five specific domains, and invite participants from five regions to record their native dialects. Additionally, we engage skilled human annotators to review all data to ensure the quality and reliability. The final multi-dialect benchmark comprises a total of 25,000 samples collected from five regions: Hunan, Minnan, Guangyue, Chuanyu, and Shanghai (each region has 5,000 cases); while the multi-domain benchmark includes 2,252 samples from five dimensions: Chat (443 cases), Government (462 cases), Health (450 cases), Knowledge (421 cases), and Local-live (476 cases).


\subsection{Quantitative Results}
\label{sec:B_5_3_performancecomp}

We conduct comprehensive evaluations of \modellite{} against state-of-the-art MLLMs on 14 different multimodal benchmarks, as illustrated in Table \ref{table_1_ocnormal}$\sim$\ref{table_10_audiotts}. Extensive experiments demonstrate that \modellite{} achieves comparable performance with leading MLLMs.

\noindent\textbf{Image $\rightarrow$ Text (Understanding)}. In addition to comparable performance on various image understanding benchmarks like grounding (Table \ref{table_4_ocobjdet}) and OCR (Table \ref{table_3_ococr}) tasks, \modellite{} also outperforms the current leading MLLMs on a series of benchmarks, particularly in GUI (Table \ref{table_3_ocgui}) and knowledge-bases QA (Table \ref{table_7_knowledge}) tasks. Moreover, \modellite{} also shows robust capabilities in addressing expert-level encyclopedia tasks (Table \ref{table_6_eyes}) and exhibits superior human-centric interaction patterns when communicating with real users (Table \ref{table_5_prefence}). 



Specifically, as illustrated in Table \ref{table_3_ocgui}, \modellite{} achieves exceptional advancements on GUI benchmarks. \modellite{} outperforms InternVL3-8B by +2.6$\%$ on ScreenSpot and +2.7$\%$ on ScreenSpot-V2. Moreover, it achieves an accuracy of 66.6$\%$ on AITZ(EM),  surpassing the reproduced results of Qwen2.5VL-7B-Insruct by +9.0$\%$, demonstrating robust GUI grounding and action reasoning capabilities in Android environment. 

In addition, \modellite{} obviously surpasses current leading MLLMs by a large margin on knowledge-intensive benchmarks. As shown in Table \ref{table_7_knowledge}, \modellite{} demonstrates superior performance compared to models even with larger parameters, achieving an obvious performance gain of +8.3$\%$/+9.8$\%$/+7.1$\%$ across three dimensions of the InfoSeek benchmark. Additionally, as presented in Table \ref{table_6_eyes}, \modellite{} achieves an average performance improvement of +5.20$\%$ over Qwen2.5VL-7B-Insruct on the in-house encyclopedia benchmark. These results indicate that \modellite{} incorporates richer expert-level knowledge and presents superior capability in querying fine-grained information from images, highlighting the effectiveness of integrating high-quality structured data and encyclopedia data.

Lastly, \modellite{} exhibits better human-centric interaction patterns when communicating with real users. As illustrated in Table \ref{table_5_prefence}, compared to Qwen2.5VL-7B-Insruct, \modellite{} presents superior capability in generating human-centric responses that are more relevant, more fluent, better formatted, and less prone to hallucination.


\noindent\textbf{Text $\rightarrow$ Image (Generation)}. As shown in Table \ref{table_g_imggen}, our experimental results demonstrate that the generation quality of \modellite{} is on par with state-of-the-art diffusion models. Notably, \model{} significantly outperforms all baselines in terms of FID, highlighting its top-performing performance in visual quality enhancement and artifact suppression. The slight drop in GenEval is attributed to a trade-off between the instruction-following ability of diffusion models and the artifact sensitivity of autoregressive models. Additionally, we observe that JanusPro achieves higher GenEval scores partly due to the use of rewritten prompts, which may not be a fair performance comparison since our results are evaluated with the original prompts. Lastly, all other baselines are tailored for basic image generation tasks with constrained generalization ability, which could hardly support the rich editing and style variation capabilities offered by our approach. This highlights the core advantage of our unified framework, which leverages the implicit understanding ability of MLLMs to enable more intelligent and controllable generation.



\noindent\textbf{Audio $\rightarrow$ Text (Understanding)}. As illustrated in Table \ref{table_8_audioasr}, \modellite{} achieves superior performance on audio understanding tasks, yielding new SoTA results on two out of seven public Chinese benchmarks and setting new SoTA on four out of six public English benchmarks (reaching a total of 6/13 SoTA). Specifically, \modellite{} outperforms Qwen2.5-Omni~\citep{xu2025qwen2} on both public audio understanding benchmarks as well as in-house multi-dialect and multi-domain testing sets in terms of average performance, demonstrating its superior capabilities in handling diverse general audio understanding tasks. In addition, \modellite{} also presents competitive performance in audio QA tasks compared to SoTA audio-centric methods and unified multi-modal ones in Table \ref{table_9_audioqa}. These results reflect the effectiveness of incorporating a high-quality audio corpus and the staged training strategies proposed in Sec. \ref{sec_2_speech_understanding}.


\noindent\textbf{Text $\rightarrow$ Audio (Generation)}. As shown in Table \ref{table_10_audiotts}, \modellite{} attains competitive results compared to SoTA text-to-speech (TTS) methods, revealing strong capabilities in handling unified audio perception-generation tasks.


\noindent\textbf{Video $\rightarrow$ Text (Understanding)}. Following the conventional protocol, we uniformly sample 128 frames for each video during evaluation. As demonstrated in Table \ref{table_11_video}, \modellite{} reaches a new SOTA in terms of the average metric across four widely-used benchmarks, outperforming Qwen2.5VL-7B-Instruct~\citep{bai2025qwen25vltechnicalreport} by +0.2$\%$ and LLaVA-OneVision-7B by +9.6$\%$ in average performance. Moreover, \modellite{} achieves a noticeable improvement of +1.9$\%$ on the LongVideoBench benchmark compared with Qwen2.5VL-7B-Instruct, revealing its superior capability in capturing and understanding informative spatial-temporal content particularly for long-duration videos.



\subsection{Visualization Results}
\label{sec:B_5_3_visualization}

To further present the unified perception and generation capabilities of \modellite{} in addressing various multi-modal tasks, we present a selection of qualitative examples through responses generated from various prompts. As illustrated in Figure \ref{part-6-figure-under-all} and \ref{part-5-fig-4-vis-gen}, \modellite{} is proficient in handling [\textit{Image, Audio, Video, Text}] $\Longleftrightarrow$ [\textit{Image, Audio, Text}] tasks. Specifically, we visualize the capabilities of \textit{Image $\rightarrow$ Text} understanding tasks, including OCR, grounding, information-seeking, and reasoning in Figure \ref{part-6-figure-under-all}. Furthermore, we also illustrate the capabilities of \textit{Video$\&$Audio $\rightarrow$ Text} understanding tasks in the bottom of the Figure \ref{part-6-figure-under-all}. Lastly, we visualize the capabilities of \textit{Text $\rightarrow$ Image} generation tasks in Figure \ref{part-5-fig-4-vis-gen}, including image generation, image editing, and style transfer.































\section{Conclusion}
\label{sec:Conclusion}

We introduce \model{}, the first open-source model we are aware of to match GPT-4o in modality support. It can perceive text, images, videos and audio modalities and generate text, natural speech in real-time and images simultaneously. \model{} is built on the Ling MoE architecture with modality-specific routers to mitigate modality conflicts, and excels in multi-modal interaction and generation. \model{} attained performance on par with Qwen2.5-VL-7B by activating only 2.8B parameters, and demonstrated SOTA performance in end-to-end speech understanding and speech instruction following.  In addition, \model{} also support native-resolution image generation, editing, and style transfer, surpasses mainstream generation models like SDXL. Overall, \model{} demonstrates robust adaptability and efficiency across multimodal perception and generation tasks, showcasing promising prospects for future research and industrial applications.
 
\newpage
\section{Contributors}
\label{sec:contri}

\large{Authors are listed \textbf{alphabetically by the first name}.}

\large{
\begin{multicols}{3}
\raggedcolumns
Ant Inclusion AI \\
Biao Gong\\
Cheng Zou\\
Chuanyang Zheng\\
Chunluan Zhou\\
Canxiang Yan \\
Chunxiang Jin \\ 
Chunjie Shen \\
Dandan Zheng \\
Fudong Wang\\
Furong Xu\\
GuangMing Yao\\      
Jun Zhou \\
Jingdong Chen \\
Jianxin Sun \\
Jiajia Liu \\
Jianjiang Zhu \\
Jun Peng\\
Kaixiang Ji\\
Kaiyou Song\\
Kaimeng Ren \\
Libin Wang\\
Lixiang Ru\\
Lele Xie \\
Longhua Tan \\
Lyuxin Xue \\ 
Lan Wang \\
Mochen Bai\\
Ning Gao \\
Pei Chen \\
Qingpei Guo \\
Qinglong Zhang \\
Qiang Xu \\
Rui Liu \\
Ruijie Xiong \\ 
Sirui Gao \\
Tinghao Liu \\
Taisong Li \\
Weilong Chai \\
Xinyu Xiao \\
Xiaomei Wang \\
Xiaoxue Chen \\
Xiao Lu \\
Xiaoyu Li \\
Xingning Dong\\
Xuzheng Yu\\
Yi Yuan \\
Yuting Gao\\
Yunxiao Sun\\
Yipeng Chen\\
Yifei Wu\\ 
Yongjie Lyu\\ 
Ziping Ma\\
Zipeng Feng\\
Zhijiang Fang\\
Zhihao Qiu\\
Ziyuan Huang\\
Zhengyu He

\end{multicols}}

\clearpage

\bibliographystyle{assets/plainnat}
\bibliography{paper}

\newpage
\beginappendix

\section{Open-source image data}
\label{sec:app_image_data}
In addition to the open source image data utilized in \citep{guo2025m2}, the newly added data are presented in Table~\ref{tab:image_data}.

\begin{table*}[bpth]
    \centering
    \small
    \caption{\centering{The list of newly added open-source image data used during our training.}}
    \begin{tabular}{c}
    \hline
         Dataset\\
         \hline
         OS-ATLAS~\citep{wu2024osatlasfoundationactionmodel}\\
         M2E~\citep{yang2023last}\\
         IM2LATEX-100K~\citep{deng2017imagetomarkupgenerationcoarsetofineattention}\\
         Mini-CASIA-CSDB~\citep{ding2022alargescaledatabaseforchemicalstructurerecognition} \\
         CASIA-CSDB~\citep{ding2022alargescaledatabaseforchemicalstructurerecognition} \\
         DoTA~\citep{liang-etal-2024-document}\\
         ICDAR23-SVRD~\citep{yu2023icdar2023competitionstructured}\\
         AitZ~\citep{zhang2024androidzoochainofactionthoughtgui} \\
         AitW~\citep{rawles2023androidwildlargescaledataset}\\
         GUICourse~\citep{wen2024guicourse} \\
         OmniMedVQA~\citep{hu2024omnimedvqanewlargescalecomprehensive}\\
         SLAKE~\citep{liu2021slakesemanticallylabeledknowledgeenhanceddataset}\\
         VQA-Med~\citep{ImageCLEF-VQA-Med2021}\\
         Geometry3K~\citep{lu2021intergpsinterpretablegeometryproblem}\\
         UniGeo~\citep{chen2022unigeounifyinggeometrylogical}\\
         MAVIS~\citep{zhang2024mavismathematicalvisualinstruction}\\
         GeoS~\citep{seo-etal-2015-solving}\\
         PixMo-count~\citep{deitke2024molmo}\\
         Geoqa+~\citep{cao-xiao-2022-augmented}\\
         GeomVerse~\citep{kazemi2023geomversesystematicevaluationlarge}\\
         ChemVLM~\citep{li2025chemvlmexploringpowermultimodal}\\

        \hline
    \end{tabular}
    \label{tab:image_data}
\end{table*}

\section{Open-source audio data}
\label{sec:app_audio_data}
We include the complete list of open source audio data we used during our training in Table~\ref{tab:audio_data}.

\begin{table*}[bpth]
    \centering
    \small
    \caption{\centering{The complete list of open-source audio data used during our training.}}
    \begin{tabular}{c|c}
    \hline
         Dataset & Audio Length (hrs) \\
         \hline
        WenetSpeech~\citep{zhang2022wenetspeech10000hoursmultidomain}	& 10518	\\
        KeSpeech~\citep{tang2021kespeech}	& 1428	\\
        AliMeeting~\citep{AliMeeting} & 120	\\
        AISHELL-1~\citep{AISHELL1}	& 155	\\
        AISHELL-3~\citep{shi2020aishell}	& 65	\\
        AISHELL-4~\citep{fu2021aishell}	& 61	\\
        CoVoST~\citep{wang2020covost}	    & 456	\\
        CoVoST2~\citep{CoVoST2}	    & 18	\\
        Magicdata~\citep{MagicData_RAMC}	& 747	\\
        Gigaspeech~\citep{GigaSpeech}	& 10288	\\
        Libriheavy~\citep{Libriheavy}	& 51448	\\
        LibriSpeech~\citep{Librispeech}	& 960	\\
        SlideSpeech~\citep{SlideSpeech}	& 473	\\
        SPGISpeech~\citep{SPGISpeech}	& 5000	\\
        TED-LIUM~\citep{rousseau2012ted}	& 208	\\
        Emilla~\citep{he2024emiliaextensivemultilingualdiverse}	    & 90305	\\
        Multilingual LibriSpeech~\citep{pratap2020mls} & 45000 \\
        Peoples Speech~\citep{galvez2021people} &	30000 \\
        \hline
    \end{tabular}
    \label{tab:audio_data}
\end{table*}

\section{Open-source video data}
\label{sec:app_video_data}

The comprehensive list of open-source video datasets utilized during the training process is provided in Table~\ref{tab:vieo_data}.

\begin{table*}[bpth]
    \centering
    \small
    \caption{\centering{The complete list of open-source video data used during our training.}}
    \begin{tabular}{c}
    \hline
         Dataset\\
         \hline
         TGIF-Transition~\citep{yuntgifqa}\\
         ShareGPT4Video~\citep{chen2024sharegpt4video}\\
         videogpt-plus~\citep{Maaz2024VideoGPT+} \\
         Llava-video-178k~\citep{llava_video}\\
         Video-Vista~\citep{li2024videovistaversatilebenchmarkvideo} \\
         Neptune~\citep{nagrani2025neptunelongorbitbenchmarking} \\
         FunQA~\citep{xie2024funqasurprisingvideocomprehension}\\
         Temp-Compass~\citep{liu2024tempcompassvideollmsreally}\\
         EgoTask~\citep{jia2022egotaskqaunderstandinghumantasks} \\
         InternVid~\citep{wang2024internvidlargescalevideotextdataset} \\
         CLEVRER~\citep{Yi*2020CLEVRER:}\\
         VLN-CE~\citep{krantz_vlnce_2020}\\
         Vript~\citep{yang2024vript}\\
         Cinepile~\citep{rawal2024cinepilelongvideoquestion}\\
         OpenVid-1M~\citep{nan2024openvid}\\

        \hline
    \end{tabular}
    \label{tab:vieo_data}
\end{table*}


\end{document}